\documentclass[runningheads]{llncs}

\usepackage{eccv}

\usepackage{eccvabbrv}

\usepackage{graphicx}
\usepackage{booktabs}
\usepackage{xcolor}
\usepackage{colortbl}
\usepackage{makecell}

\usepackage[accsupp]{axessibility}  

\usepackage{hyperref}

\usepackage{orcidlink}

\begin{document}

\title{Asynchronous Large Language Model Enhanced Planner for Autonomous Driving} 

\titlerunning{Asynchronous LLM Enhanced Planner for Autonomous Driving}

\author{Yuan Chen\inst{1,2}\thanks{Equal contribution. $^\dagger$ Corresponding author. }\and
Zi-han Ding\inst{1}$^\star$ \and
Ziqin Wang\inst{1}$^\star$ \and
Yan Wang\inst{2}$^\dagger$ \and
Lijun Zhang\inst{1} \and
Si Liu\inst{1}$^\dagger$
}

\authorrunning{Y. Chen et al.}

\institute{Beihang University \and AIR, Tsinghua University \\
\email{\{chenyuan1, wzqin, ljzhang, liusi\}@buaa.edu.cn} \\
\email{zihanding819@gmail.com} \\
\email{wangyan@air.tsinghua.edu.cn}}

\maketitle

\begin{abstract}

Despite real-time planners exhibiting remarkable performance in autonomous driving, the growing exploration of Large Language Models (LLMs) has opened avenues for enhancing the interpretability and controllability of motion planning. Nevertheless, LLM-based planners continue to encounter significant challenges, including elevated resource consumption and extended inference times, which pose substantial obstacles to practical deployment. In light of these challenges, we introduce \textit{AsyncDriver}, a new asynchronous LLM-enhanced closed-loop framework designed to leverage scene-associated instruction features produced by LLM to guide real-time planners in making precise and controllable trajectory predictions. On one hand, our method highlights the prowess of LLMs in comprehending and reasoning with vectorized scene data and a series of routing instructions, demonstrating its effective assistance to real-time planners. On the other hand, the proposed framework decouples the inference processes of the LLM and real-time planners. By capitalizing on the asynchronous nature of their inference frequencies, our approach have successfully reduced the computational cost introduced by LLM, while maintaining comparable performance. Experiments show that our approach achieves superior closed-loop evaluation performance on nuPlan's challenging scenarios. The code and dataset are available at \url{https://github.com/memberRE/AsyncDriver}.

\end{abstract}

\section{Introduction}
\label{sec:intro}

\begin{figure}[tbp]
    \centering
    \includegraphics[width=1\textwidth]{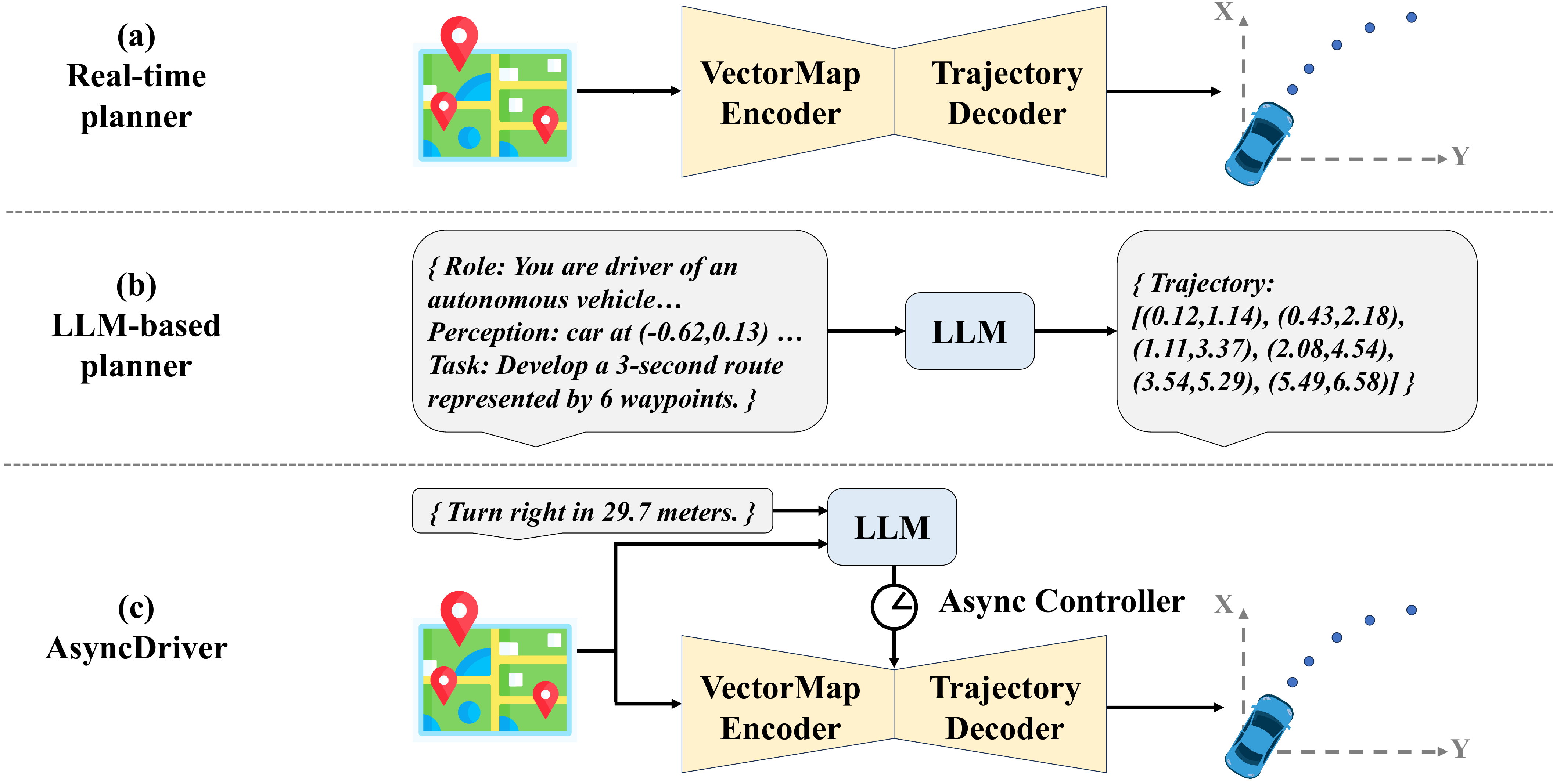}
    \caption{\textbf{Comparative Overview of Learning-based Autonomous Driving Planning Frameworks.} (a) Real-time planner: Offers quick inference but has limited controllability. (b) LLM-based planner: Produces linguistic descriptions and controls, offering high interactivity and interpretability at the expense of inference speed. (c) AsyncDriver: While leveraging the reasoning capabilities of LLM, a balance between performance and inference speed is achieved through asynchronous control.}
    \label{fig:1}
\end{figure}

Motion planning plays a pivotal role in autonomous driving, garnering significant interest due to its direct impact on vehicle navigation and safety. One particularly noteworthy evaluation approach is the employment of closed-loop simulation, which involves the dynamic development of driving scenarios that adapt to the planner's predicted trajectories, thus necessitating that the model exhibits stronger predictive accuracy and bias correction capabilities.

As illustrated in Fig.\ref{fig:1}(a), current learning-based real-time motion planning frameworks \cite{kendall2019learning,li2023boosting,hallgarten2023gcpgp,renz2022plant,scheel2022urban,huang2023gameformer} typically utilize vectorized map information as input and employ a decoder to predict trajectories. As purely data-driven methods, they are particularly vulnerable to long-tail phenomena, where their performance can significantly degrade in rare or unseen scenarios \cite{chen2022milestones}. Moreover, while some rule-based strategies exist, their manual crafting of rules are found to be inadequate for capturing the entirety of potential complex scenarios, resulting in driving strategies that tend towards extremes either excessive caution or aggression. Furthermore, both learning-based and rule-based planning frameworks suffer from low controllability, which raises concerns regarding the safety and reliability of such systems in dynamic environments.

Recently, the considerable potential of Large Language Models (LLMs), including GPT-4 \cite{achiam2023gpt} and Llama2 \cite{touvron2023llama}, has been extensively explored within the realm of autonomous driving. Their extensive pre-training on large-scale datasets has established a robust foundation for comprehending traffic rules and scenarios. Consequently, LLM-based planners have demonstrated superior performance in scene analysis, reasoning, and human interaction, heralding new prospects for enhancing the interpretability and controllability of motion planning \cite{zhou2023vision,yang2023survey,cui2024survey}. Nonetheless, as shown in Fig.\ref{fig:1}(b), these models frequently encounter several of these specific challenges: 1) The scene information is described through language, which could be constrained by the permissible input token length, making it challenging to encapsulate complex scene details comprehensively and accurately \cite{mao2023gpt,sha2023languagempc,wen2023dilu,mao2023language}.  2) Prediction via linguistic outputs entails either directly issuing high-level commands that are then translated into control signals, potentially leading to inaccuracies, or outputting trajectory points as floating-point numbers through language, a task at which LLMs are not adept \cite{xu2023drivegpt4,mao2023language,keysan2023can}. 3) Prevalent frameworks primarily utilize LLMs as the core decision-making entity. While this strategy offers advantages in performance, the inherently large number of parameters in LLMs results in a noticeable decrease in inference speed relative to real-time planners, presenting substantial obstacles to their real-world implementation.

In this work, we introduce AsyncDriver, a novel asynchronous LLM-enhanced framework for closed-loop motion planning. As depicted in Fig.\ref{fig:1}(c), this method aligns the modalities of vectorized scene information and series of routing instructions, fully leveraging the considerable capabilities of LLM for interpreting instructions and understanding complex scenarios. The Scene-Associated Instruction Feature Extraction Module extracts high-level instruction features, which are then integrated into the real-time planner through the proposed Adaptive Injection Block, significantly boosting prediction accuracy and ensuring finer trajectory control. Moreover, our approach preserves the architecture of the real-time planner, allowing for the decoupling of inference frequency between LLM and the real-time planner. By controlling the asynchronous intervals of inference, it significantly enhances computational efficiency and alleviates the additional computational cost introduced by LLM. Furthermore, the wide applicability of our proposed Adaptive Injection Block ensures that our framework can be seamlessly extended to any transformer-based real-time planner, underscoring its versatility and potential for broader application.

To summarize, our paper makes the following contributions:
\begin{itemize}
    \item We propose AsyncDriver, a novel asynchronous LLM-enhanced framework, in which the inference frequency of LLM is controllable and can be decoupled from that of the real-time planner. While maintaining high performance, it significantly reduces the computational cost.
    \item We introduce the Adaptive Injection Block, which is model-agnostic and can easily integrate scene-associated instruction features into any transformer-based real-time planner, enhancing its ability in comprehending and following series of language-based routing instructions.
    \item Compared with existing methods, our approach demonstrates superior closed-loop evaluation performance in nuPlan's challenging scenarios.
\end{itemize}

\section{Related Work}
\label{sec:related_work}

\subsection{Motion Planning For Autonomous Driving}
The classic modular pipeline for autonomous driving includes perception, prediction, and planning. In this framework, the planning stage predict a future trajectory based on the perception outputs, then executed by the control system. This architecture, widely adopted in industry frameworks like Apollo \cite{apollo_auto}, contrasts with end-to-end approaches \cite{hu2023planning,hu2022st} by enabling focused research on individual tasks through well-defined data interfaces between modules.

Autonomous driving planners can be mainly categorized into rule-based and learning-based types. Rule-based planners \cite{idm_2002,Kesting_Treiber_Helbing_2007,pdm_2023,fan2018baidu,urmson2008autonomous,leonard2008perception,bacha2008odin} rely on predefined rules for determining the vehicle's trajectory, such as maintaining a safe following distance and obeying traffic signals. For instance, IDM \cite{idm_2002} ensures a safe distance from the leading vehicle by calculating an appropriate speed based on braking and stopping distances. PDM \cite{pdm_2023} builds on IDM by selecting the highest-scoring IDM proposal as the final trajectory, achieving state-of-the-art performance in the nuPlan Challenge 2023 \cite{caesar2021nuplan}. However, rule-based planners often struggle with complex driving scenarios beyond their predefined rules.

Learning-based planners \cite{kendall2019learning,li2023boosting,hallgarten2023gcpgp,renz2022plant,scheel2022urban,huang2023gameformer} aim to replicate human expert driving trajectories using imitation learning or offline reinforcement learning from large-scale datasets. However, they face limitations due to the scope of the datasets and model complexity, leaving substantial room for improvement in areas like routing information comprehension and environmental awareness.

\subsection{LLM For Autonomous Driving}
The rapid advancement of Large Language Models has been noteworthy. These models, including GPT-4 \cite{achiam2023gpt} and Llama2 \cite{touvron2023llama}, have been trained on extensive textual datasets and exhibit exceptional generalization and reasoning abilities. A growing body of research has explored the application of LLMs' decision-making capacities to the domain of autonomous driving planning \cite{sima2023drivelm,chen2023driving,mao2023gpt,sha2023languagempc,jin2023surrealdriver,wen2023dilu,mao2023language,wang2023drivemlm,shao2023lmdrive,sharan2023llm,han2024dme,xu2023drivegpt4,fu2024drive,cui2023receive,wang2023chatgpt,ma2023dolphins,nie2023reason2drive,ma2023lampilot,chen2023towards,liu2023mtd,yuan2024rag,wang2023empowering}.

Some efforts \cite{fu2024drive,mao2023language,wen2023dilu,mao2023gpt} have explored integrating scene information, including the ego vehicle's status and information about obstacles, pedestrians, and other vehicles into LLMs using linguistic modalities for decision-making and explanations. These approaches face limitations due to finite context length, making it challenging to encode precise information for effective decision-making and reasoning. To overcome these constraints, multi-modal strategies such as DrivingWithLLM \cite{chen2023driving}, DriveGPT4 \cite{xu2023drivegpt4}, and RAGDriver \cite{yuan2024rag} have been developed. These methods align vectorized or image/video modalities with linguistic instructions for a more comprehensive interpretation of driving scenarios. However, using language to express control signals has its limitations. DrivingWithLLM outputs high-level commands in linguistic expressions, improving QA interaction but reducing the fidelity of translating complex reasoning into precise vehicle control. DriveGPT4 expresses waypoints through language, showing strong open-loop performance but lacking closed-loop simulation evaluation.

Furthermore, some efforts \cite{sha2023languagempc,shao2023lmdrive} focus on closed-loop evaluation by connecting low-level controllers or regressors behind large language models for precise vehicle control. LMDrive \cite{shao2023lmdrive} uses continuous image frames and navigation instructions for closed-loop driving but requires complete LLM inference at each planning step. LanguageMPC \cite{sha2023languagempc} employs LLMs to obtain Model Predictive Control parameters, achieving control without training. However, these methods necessitate serial language decoding or full LLM inference at each planning step, challenging real-time responsiveness and limiting practical deployment.

In our approach, we shift from using LLMs for direct language output to enhancing real-time learning-based planners. This strategy improves environmental comprehension and allows LLMs and real-time planners to operate independently at different inference rates. This decoupling reduces LLM inference latency, facilitating real-world deployment.

\section{Data Generation}
\label{sec:datageneration}
The nuPlan \cite{caesar2021nuplan} dataset is the first large-scale benchmark for autonomous driving planning, comprising $1,200$ hours of real-world human driving data from Boston, Pittsburgh, Las Vegas, and Singapore. To support various training stages, we developed pre-training and fine-tuning datasets from the nuPlan \textit{Train} and \textit{Val} sets, focusing on 14 official \cite{nuplanchallange} challenging scenario types.

\subsection{Pre-training Data Generation}
To enhance LLM's understanding of instructions in autonomous driving, we created a dataset of language-based QAs, aligning with LLM's native modality to better grasp instruction semantics, which includes \textit{Planning-QA} and \textit{Reasoning1K}, with sample datasets provided in the supplementary material.

\subsubsection{Planning-QA}

is created using a rule-based approach for scalability. It is designed to enhance the LLM's understanding of the relationships among waypoints, high-level instructions, and control. In this context, waypoints are arrays of points, high-level instructions are composed of velocity commands (\textit{stop, accelerate, decelerate, maintain speed}) and routing commands (\textit{turn left, turn right, go straight}), and control involves velocity and acceleration values. Planning-QA includes six types of questions, each focusing on the conversion between waypoints, high-level instructions, and control.

\subsubsection{Reasoning1K}
includes $1,000$ pieces of data generated by GPT-4, beyond merely providing answers, it further supplements the reasoning and explanation based on the scene description and is used for mixed training with Planning-QA.

\subsection{Fine-tuning Data Generation}

To further achieve multimodal understanding and alignment, we constructed a fine-tuning dataset based on $10,000$ scenarios, capturing one frame every 8 seconds, resulted in a training set of $180,000$ frames and a validation set of $20,000$ frames, each incorporating both vectorized map data and linguistic prompts. Importantly, the scenario type distribution in both training and validation datasets matches the distribution of the entire nuPlan trainval dataset.

For the extracted vectorized scene information, similar to \cite{huang2023gameformer}, ego information and $20$ surrounding agent information over $20$ historical frames, in addition to global map data centered around the ego are involved. 

LLM's prompt is comprised of two parts: system prompt and series of routing instructions. 
Concerning routing instructions, a rule-based approach is employed to transform the pathway into a series of instructions enhanced with distance information.
Regarding training dataset preparation, the ego vehicle's ground truth trajectory over the ensuing $8$ seconds is harnessed as the pathway for routing instruction generation. During simulation, based on the observation of the current scene, a pathway that adheres to a specified maximum length is found through a hand-crafted method as a reference path for instruction generation.

\section{Methodology}
\label{sec:meth}

\begin{figure}[tbp]
    \centering
    \includegraphics[width=1\textwidth]{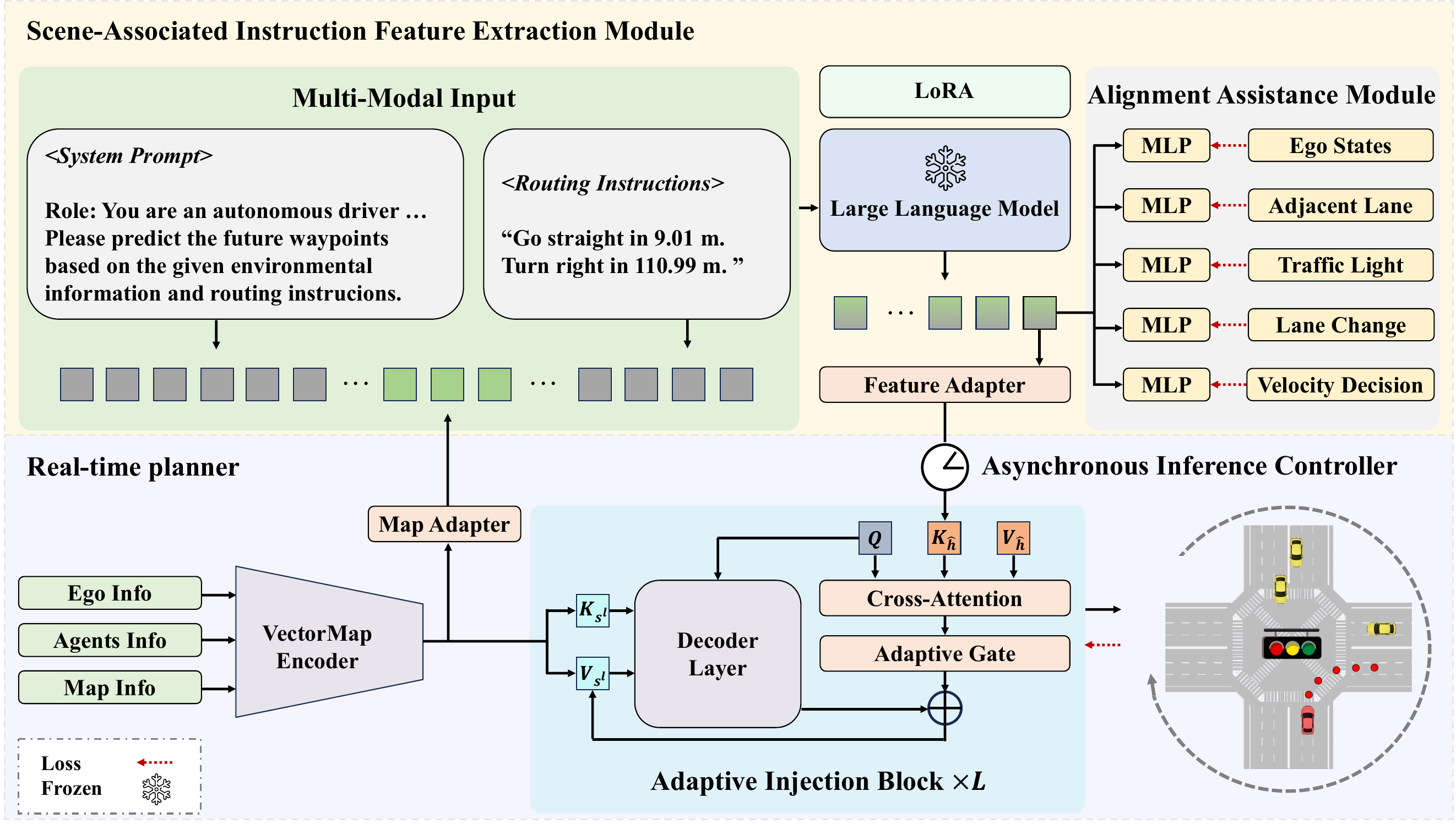}
    \caption{\textbf{Overview of our proposed AsyncDriver framework.} Scene information, together with routing instructions, is encoded through the Scene-Associated Instruction Feature Extraction Module. Subsequently, the Adaptive Injection Block asynchronously enhances the features of the real-time planner, facilitating closed-loop control for autonomous vehicles. The Alignment Assistance Module is exclusively employed for multi-modality alignment during training.}
    \label{fig:pipeline}
\end{figure}

As illustrated in Fig. \ref{fig:pipeline}, we introduce the asynchronous LLM-enhanced closed-loop framework, AsyncDriver, which mainly includes two components: 1) The Scene-Associated Instruction Feature Extraction Module; 2) The Adaptive Injection Block. Additionally, due to our framework’s design, the inference frequency between the LLM and the real-time planner could be decoupled and regulating by asynchronous interval, which markedly enhancing inference efficiency.

Within this section, we present the Scene-Associated Instruction Feature Extraction Module (Section \ref{sec:41}), detail the design of Adaptive Injection Block (Section \ref{sec:42}), discuss the concept of Asynchronous Inference (Section \ref{sec:43}) and outline the training details employed (Section \ref{sec:44}).

\subsection{Scene-Associated Instruction Feature Extraction Module}
\label{sec:41}

\subsubsection{Multi-modal Input}

In each planning iteration, the vectorized scene information is acquired from the simulation environment. Analogous to the approach employed by GameFormer \cite{huang2023gameformer}, historical trajectory and state information of both the ego and other agents are extracted, alongside global map data. Vectorized scene information for the real-time planner is provided in the same manner. All vector data are relative to the position of the ego. Subsequently, through the processing by the Vector Map Encoder and Map Adapter, we derive map embeddings. These map embeddings, along with language embeddings, are then fed into the Llama2-13B backbone to obtain the final hidden features $h = \{h_0, h_1, ..., h_{-1}\} \in \mathbb{R}^{N_h \times D_{llm} }$.

\subsubsection{Alignment Assistance Module}
To grasp the essence of routing instructions while maintaining a fine-grained comprehension of vectorized scene information for enhanced extraction of scene-associated high-level instruction features, we employ the Alignment Assistance Module to facilitate the alignment of multi-modal input. Concretely, we have pinpointed five critical scene elements essential to the autonomous driving process for multi-task prediction, which is implemented by five separate 2-layer MLP prediction heads. About the current states of the ego vehicle, we perform regression to estimate the vehicle's velocity and acceleration along the X and Y axes. For map information, we undertake classification tasks to identify the existence of adjacent lanes on both the left and right sides and to assess the status of traffic lights relevant to the current lane. Additionally, with a view towards future navigation strategies, we classify the requirement for lane changes in upcoming trajectories and identify future velocity decision, which includes options \textit{acceleration}, \textit{deceleration}, and \textit{maintaining current speed}. It is worth noting that the Alignment Assistance Module is only used to assist multi-modal alignment in the training phase and does not participate in the inference stage. 

\subsection{Adaptive Injection Block}
\label{sec:42}
We adopt the decoder structure of \cite{huang2023gameformer} as our basic decoder layer, facilitating the adaptive integration of scene-associated instruction features by evolving the conventional transformer-based decoder layer into an Adaptive Injection Block.

Specifically, the hidden feature of the last token ${h}_{-1}$ is projected via the feature adapter and subsequently fed into Adaptive Injection Block.

\begin{equation}
\hat{h}=Linear({h}_{-1})
\end{equation}

Within the Adaptive Decoder Block, the foundational decoder architecture of the real-time planner is elegantly extended to ensure that the query in each layer not only preserves the attention operation intrinsic to the original scene information but also engages in cross-attention with scene-associated instruction features, thereby incorporating instructional guidance into the prediction process. Afterwards, the updated instruction-enhanced query feature is modulated by the learnable adaptive gate, which is initialized by zero and reintegrated into the original attention output of the decoder layer. The adaptive injection process of the \textit{l-th} decoder block can be formulated as follows:

\begin{equation}
s^{l+1}={g\cdot softmax({\frac{Q{{K}_{\hat{h}}}^T}{\sqrt{C}}})V_{\hat{h}}+softmax({\frac{Q{{K_{s^{l}}}}^T}{\sqrt{C}}})V_{s^{l}}}
\end{equation}
where $g$ is the value of adaptive gate, $Q$ denotes query in original decoder layer, $K_{i}$ and $V_{i}$ represent key and value respectively of feature $i$, and $s^{l}$ notes the scene feature of the \textit{l-th} decoder layer.

The proposed adaptive injection method not only maintains the original decoder layer's ability to process complete scene information within the real-time planner but also enhances the planner's understanding and compliance with a series of flexible linguistic instructions. This advancement allows for the production of more refined and controllable predictions.
It is worth noting that due to the simple yet effective design of our Adaptive Injection Block, it can be seamlessly integrated into any transformer-based architecture, thereby affording our approach the flexibility to be adapted to other real-time planner frameworks.

\subsection{Asynchronous Inference}
\label{sec:43}
Our design leverages LLM to guide the real-time planner, significantly enhancing its performance through series of flexibly combined linguistic instructions without compromising its structural integrity. This method facilitates controlled asynchronous inference, effectively decoupling the inference frequencies of the LLM and the real-time planner, thereby LLM is not required to process every frame. During asynchronous intervals, the previously derived high-level instruction features continue to guide the prediction process of the real-time planner, which significantly boosts the inference efficiency and reduces the computational cost introduced by LLM. Notably, our framework accommodates a series of flexibly combined routing instructions, capable of delivering long-term, high-level routing insights. Therefore, even amidst asynchronous intervals, prior high-level features could still offer effective guidance, reinforcing the performance robustness throughout LLM inference intervals.

Experimental results reveal that our architecture maintains near-robust performance when the asynchronous inference interval of LLM is extended. By controlling the LLM to perform inference every $3$ frames can achieve a reduction in inference time of nearly $40\%$, with only a minimal accuracy loss of about $1\%$, which demonstrates the efficacy of our approach in striking an optimal balance between accuracy and inference speed. For a more comprehensive exploration of experimental results and their analysis, please refer to Section \ref{sec:async}.

\subsection{Training Details}
\label{sec:44}
During the pre-training stage, the entirety of Reasoning1K, augmented with $1,500$ samples randomly selected from Planning-QA, was utilized to train LoRA. This process enabled the LLM to evolve from a general-purpose large language model into one specifically optimized for autonomous driving. As a direct outcome of this focused adaptation, the LLM has become adept at understanding instructions more accurately within the context of motion planning.

During the fine-tuning stage, since the architectures of VectorMap Encoder and Decoder are preserved, we load weights of the real-time planner pre-trained on the same dataset to enhance training stability. The total loss of the fine-tuning stage is comprised of Alignment Assistance Loss and Planning Loss. The former is partitioned into five components: $l1$ loss for 1) ego velocity and acceleration prediction $\tilde{{x}}_{va} \in \mathbb{R}^4$, cross-entropy loss for 2) velocity decision prediction $\tilde{{x}}_{dec} \in \mathbb{R}^3$ and 3) traffic light state prediction $\tilde{{x}}_{traf} \in \mathbb{R}^4$, binary cross-entropy loss for 4) adjacent lane presence prediction $\tilde{{x}}_{adj} \in \mathbb{R}^2$ and 5) lane change prediction $\tilde{{x}}_{chg} \in \mathbb{R}$. The complete  Alignment Assistance Loss can be expressed as follows, where $\tilde{x}$ and ${x}$ represent prediction and ground truth respectively:

\begin{align}
{L}_{align}={L}_{1}(\tilde{{x}}_{va},{x}_{va})+CE(\tilde{{x}}_{dec},{x}_{dec})+CE(\tilde{{x}}_{traf},{x}_{traf}) \notag \\
+BCE(\tilde{{x}}_{adj},{x}_{adj})+BCE(\tilde{{x}}_{chg},{x}_{chg})
\end{align}

Following \cite{huang2023gameformer}, the planning loss comprises two parts: 1) Mode prediction loss. $m$ different modes of trajectories of neighbor agents are represented by Gaussian mixture model (GMM). For each mode, at any given timestamp $t$, its characteristics are delineated by a mean ${\mu}^{t}$ and covariance ${{\sigma}^{t}}$ forming a Gaussian distribution. The best mode ${m}^{*}$ is identified through alignment with the ground truth and refined via the minimization of negative log-likelihood. 2) Ego Trajectory Prediction loss. Future trajectory points of ego vehicle are predicted and are refined by $l1$ loss. Consequently, the planning loss is as follows:
\begin{align}
{L}_{plan} = \sum_{t}{L}_{NLL}(\tilde{\mu}_{{m}^{*}}^{t},\tilde{\sigma}_{{m}^{*}}^{t}, \tilde{p}_{{m}^{*}}, \tilde{s}_{t})+\sum_{t}{L}_{1}({\tilde{{s}}_{t}-{s}_{t})}
\end{align}

where $\tilde{\mu}_{{m}^{*}}^{t}$, $\tilde{\sigma}_{{m}^{*}}^{t}$, $\tilde{p}_{{m}^{*}}$, $\tilde{s}_{t}$ represents the predicted mean, covariance, probability, and position respectively, corresponding to the best mode ${m}^{*}$ at timestamp $t$, and ${s}_{t}$ indicates the ground truth position at timestamp $t$.

In summary, the complete loss of fine-tuning stage is formulated as:

\begin{align}
{L} = {L}_{align} + {L}_{plan}
\end{align}

\section{Experiments}
\label{sec:meth}

\subsection{Experimental Setup}

\subsubsection{Evaluation Settings}

In accordance with the nuPlan challenge $2023$ \cite{nuplanchallange} settings, we selected $14$ official challenging scenario types for training and evaluation. While nuPlan \cite{caesar2021nuplan} includes $757,844$ scenarios, most simple scenarios are insufficient for critical planner performance assessment, and the vast data volume extends evaluation times. We chose the \textit{Hard20} dataset by randomly selecting $100$ scenarios per type from the test set, then using the PDM \cite{pdm_2023} planner (nuPlan $2023$ champion) to retain the $20$ lowest-scoring scenarios per type, resulting in a test set of $279$ scenarios.
\begin{table}[tb]
  \caption{\textbf{Evaluation on nuPlan Closed-Loop Reactive Challenges on \textit{Hard20} split}. The best results are highlighted in \textbf{bold} , while the second-best results are underscored with an \underline{underline} for clear distinction. \textit{Score}: final score in average. \textit{Drivable}: drivable area compliance. \textit{Direct.}: driving direction compliance. \textit{Comf.}: ego is comfortable. \textit{Prog.}: ego progress along expert route. \textit{Coll.}: no ego at-fault collisions. \textit{Lim.}: speed limit compliance. \textit{TTC}: time to collision within bound.
  }
  
  \centering
  \setlength{\tabcolsep}{3.5pt}
  \setlength{\arrayrulewidth}{0.2pt}
\begin{tabular}{c >{\columncolor{lightgray!50}}c ccccccc}
    \toprule
    Method                              & Score & Drivable & Direct. & Comf. & Prog. & Coll. & Lim. & TTC   \\
    \midrule
    UrbanDriver \cite{scheel2022urban}                         & 35.35 & 75.53    & 97.12     & \textbf{98.56}    & \textbf{85.23}   & 55.21      & 81.62       & 47.84 \\
    GCPGP \cite{hallgarten2023gcpgp}                              & 36.85 & 81.29    & 98.20      & 77.33   & 46.96    & 72.30       & 97.92       & 68.34 \\
    IDM \cite{idm_2002}                               & 53.07 & 84.94    & 98.02     & 83.15   & 64.79    & 74.01      & 96.38       & 60.57 \\
    GameFormer \cite{huang2023gameformer}                          & 62.05 & 93.54    & 98.74     & 83.15   & 66.27    & 86.02     & 98.19       & \underline{74.55} \\
    PDM-Hybird \cite{pdm_2023}                         & 64.05 & 95.34    & 99.10      & 75.98   & 67.93    & \textbf{87.81}       & \textbf{99.57}        & 72.75 \\
    PDM-Closed \cite{pdm_2023}                         & 64.18 & \underline{95.69}    & \underline{99.10}       & 77.06   & \underline{68.20}     & \textbf{87.81}       & \textbf{99.57}        & 73.47 \\
    \midrule
    AsyncDriver                        & \underline{65.00}  & 94.62 & 98.75 & 83.15 & 67.13 & 85.13 & 98.15 & 73.48 \\ 
    AsyncDriver*                  & \textbf{67.48}  & \textbf{96.77}   & \textbf{99.10}   & \underline{83.87}   & 66.30     & \underline{87.63}     & \underline{98.24}       & \textbf{76.70}   \\
  \bottomrule
  \end{tabular}
  \label{tab:mainres}
\end{table}

\begin{table}[tp]
  \caption{
  \textbf{Scores per scenario types in evaluation on nuPlan Closed-Loop Reactive Challenges on \textit{Hard20} split}. The best results are highlighted in \textbf{bold} , while the second-best results are underscored with an \underline{underline} for clear distinction. Types $0$-$13$ represent the $14$ official scenario types of the nuPlan challenge 2023 \cite{nuplanchallange}, with specific details provided in the supplementary materials.
  }

  \label{tab:per_senario}
  \centering
  \setlength{\tabcolsep}{3.5pt}
  \setlength{\arrayrulewidth}{0.2pt}
\resizebox{\linewidth}{!}{\small
\begin{tabular}{ccccccccccccccc}
    \toprule
Methods                             & type0 & type1 & type2 & type3 & type4 & type5 & type6 & type7 & type8 & type9 & type10 & type11 & type12 & type13 \\
    \midrule
UrbanDriver \cite{scheel2022urban}                        & 69.39 & 15.78 & 44.59 & 7.42  & 13.7  & 27.14 & 0.00     & 19.44 & 20.8  & 23.61 & 68.33  & 93.16  & 33.78  & 56.39  \\
GCPGP \cite{hallgarten2023gcpgp}                              & 59.50 & 35.91 & 33.6  & 29.33 & 42.84 & 17.24 & 4.86  & 32.33 & 8.22  & 36.23 & 71.32  & 80.46  & 36.19  & 23.61  \\
IDM \cite{idm_2002}                                & 70.69 & 44.84 & \underline{91.54}  & 54.08 & 50.22 & 41.71 & 4.76  & 53.97 & 37.53 & 60.33 & 83.97  & 93.03  & 45.77  & 12.44  \\
GameFormer \cite{huang2023gameformer}                         & 84.32 & 65.78 & 83.62 & 49.03 & 71.79 & 36.8  & 0.00     & 51.76 & 55.03 & 77.24 & 82.83  & 98.24  & 49.41  & 56.16  \\
PDM-Hybird \cite{pdm_2023}                         & \textbf{87.68}  & 69.61 & 87.20  & 49.95 & 82.80  & 41.32 & 4.23  & 54.98 & 51.72 & \underline{82.95}  & 80.37  & \underline{98.40}    & 37.77  & \underline{71.99}   \\
PDM-Closed \cite{pdm_2023}                         & \underline{87.67} & \underline{70.93} & 87.20  & 49.95 & 82.80  & 41.25 & 4.22  & 53.53 & 51.57 & \textbf{82.96} & 80.37  & \textbf{98.40} & 39.95  & \textbf{72.04}  \\
\midrule
AsyncDriver                         & 83.53 & 67.24 & 83.14 & \textbf{62.89} & \underline{83.28}  & \underline{49.26}  & \underline{5.94}   & \underline{62.53}  & \textbf{65.96} & 64.86 & \textbf{84.88} & 96.62 & \underline{49.60}  & 54.35 \\ 
AsyncDriver*                        & 83.12 & \textbf{74.04}  & \textbf{91.82} & \underline{62.42} & \textbf{85.26} & \textbf{53.51} & \textbf{6.01}  & \textbf{65.26} & \underline{57.31}  & 77.16 & \underline{84.02}  & 97.56  & \textbf{62.18}  & 49.32  \\
  \bottomrule
  \end{tabular}}
\end{table}

\subsubsection{Implementation Details}
Regarding implementation details, all experiments were conducted in the closed-loop reactive setting, where agents in scenarios could be equipped with an IDM \cite{idm_2002} Planner, enabling them to react to ego vehicle's maneuvers. The simulation frequency is $10Hz$, at each iteration, the predicted trajectory has a time horizon of $8s$. We follow the closed-loop metrics proposed in nuPlan challenge, which is detailed in the supplementary material. For model settings, our AsyncDriver is built based on Llama2-13B \cite{touvron2023llama}, LoRA \cite{hu2021lora} was configured with $R=8$ and $\alpha=32$. We use the AdamW optimizer and warm-up with decay scheduler with learning rate $0.0001$.

\begin{figure*}[htbp]

    \centering
    \captionsetup[subfloat]{font=scriptsize, labelfont=normalfont, textfont=normalfont}
    \subfloat[score variation with intervals]{
        \includegraphics[height=0.24\textheight,trim=22 5 5 10,clip]{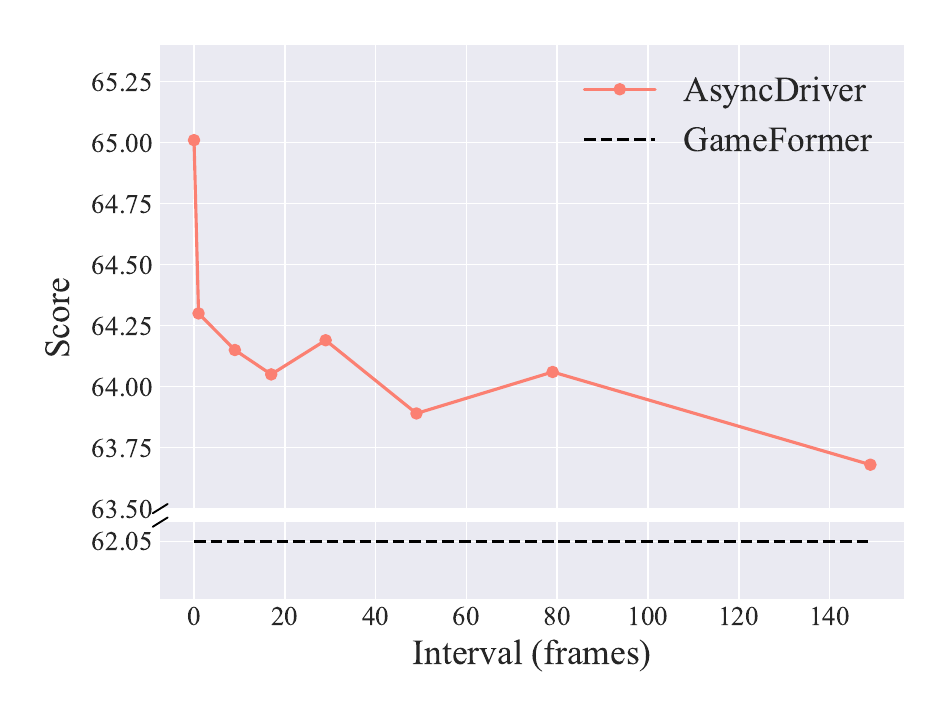}
    }
    \subfloat[average inference time variation with intervals]{
        \includegraphics[height=0.24\textheight,trim=12 5 5 10,clip]{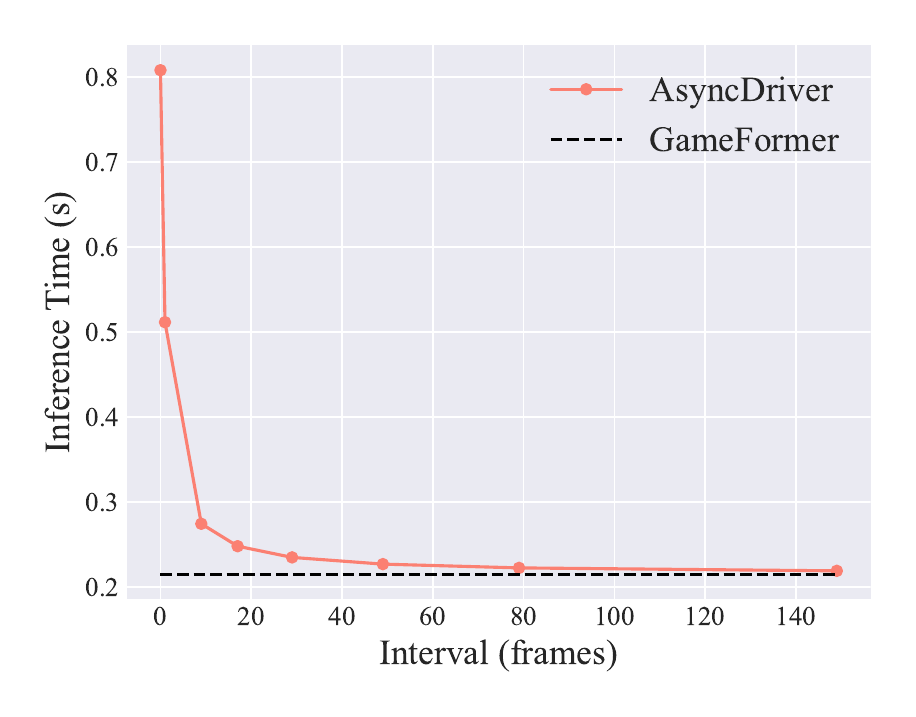}
    }
    \caption{\textbf{Asynchronous Inference.} Evaluation of expanding the inference interval to $[1,9,17,29,49,79,149]$ between the LLM and the real-time planner, executing asynchronous inference. Inference time measured on GPU Tesla A30.}

    \label{fig:decoup}
\end{figure*}

\subsection{Main Results}
\subsubsection{Hard20 Evaluation}
As illustrated in Tab. \ref{tab:mainres},  our approach AsyncDriver achieved the highest performance on \textit{Hard20} compared with existing planners, leading to an improvement of $4.6\%$ over GameFormer \cite{huang2023gameformer} about $2.95$ in score, even surpassing the current SOTA rule-based planners. For fair comparison, given the significant impact of trajectory refinement and alignment in closed-loop evaluations, we adapt the PDM \cite{pdm_2023} scorer to our AsyncDriver (denoted as AsyncDriver*), which lead to an improvement of $5.3\%$ and $5.1\%$ over PDM-Hybrid \cite{pdm_2023} and PDM-Closed \cite{pdm_2023} respectively, equivalent to approximately $3.43$ and $3.30$ in score, and an $8.7\%$ increase (approximately $5.43$ in score) over the learning-based planner GameFormer. From a different perspective, Tab. \ref{tab:per_senario} illustrates the scores of each individual scenario type on the \textit{hard20} split, as well as a comparison with existing planners. It is evident that our solution has delivered exceptional results in the majority of scenario types.

Quantitative results show that the high-level features extracted by our Scene-Associated Instruction Feature Extraction Module significantly enhance real-time planners' performance in closed-loop evaluation. Detailed metrics reveal that our approach improves drivable area performance by $3.23$ points compared to GameFormer, demonstrating superior capability in identifying and navigating viable driving spaces due to advanced scene contextual understanding. Additionally, AsyncDriver* outperforms PDM in Time to Collision (TTC) metrics by $4.39$\%, approximately $3.23$ points, indicating enhanced predictive accuracy, which is crucial for ensuring safer driving by effectively forecasting and reducing the potential collision scenarios.

\subsubsection{Asynchronous Inference}
\label{sec:async}
We contend that, particularly for generalized high-level instructions, there exhibits a notable similarity within frames of short intervals. Consequently, given its role in extracting these high-level features, the LLM does not require involvement in the inference process for each frame, which could markedly enhance the inference speed.
To explore this, experiments were designed to differentiate the inference frequency of the LLM and real-time planners, and during each LLM inference interval,  the previous instruction features are employed to guide the predicted process of the real-time planner.
As depicted in Fig. \ref{fig:decoup}, the performance of our approach demonstrates remarkable robustness as the planning interval of the LLM increases, which suggests that LLM is capable of providing a long-term high-level instructions. We observed that even with an interval of $149$ frames, meaning only one inference is made within a scenario, it still surpasses GameFormer by more than $1.0$ point, while the inference time is nearly at the real-time level. As the inference interval increases, the required inference time drops dramatically, while accuracy remains almost stable. Therefore, by employing a strategy of dense training with asynchronous inference, our method achieves an optimal balance between accuracy and inference speed. 

\subsubsection{Instruction Following}
Fig. \ref{fig:stop} shows the reactions of our method to different routing instructions, demonstrating its capability in instruction following. Fig. \ref{fig:stop_1} illustrates the outcomes predicted when the scene employs conventional routing instructions. We note that the ego vehicle decelerates slightly, a maneuver that reflects the common sense of slowing down for curves. Nonetheless, given the open road conditions, the ego vehicle sustains a comparatively high velocity. In contrast, Fig. \ref{fig:stop_2} depicts the scenario where \textit{stop} serves as the routing instruction. Remarkably, even without the presence of external obstacles, the ego vehicle promptly executes a braking response to the command, reducing its speed from $10.65 m/s$ to $1.06 m/s$ within a mere $6$ seconds. Consequently, it is evident that our AsyncDriver can function as a linguistic interaction interface, providing the ability of precisely interpreting and following human instructions to circumvent anomalous conditions.

\begin{figure}[htbp]

    \centering
\captionsetup[subfloat]{font=scriptsize, labelfont=normalfont, textfont=normalfont}
    \subfloat[receiving conventional routing instructions]{
        \includegraphics[width=1\textwidth]{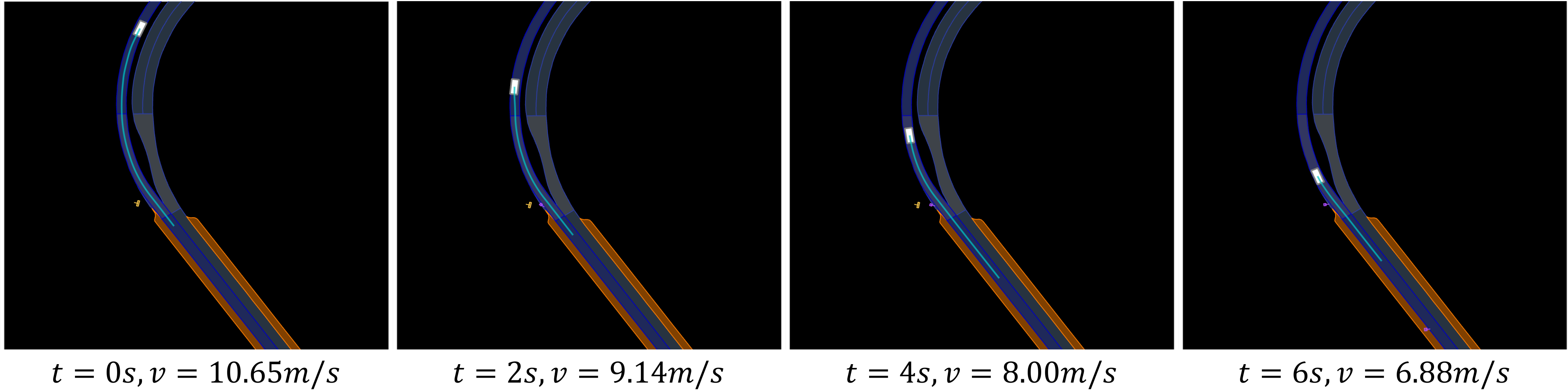}
        \label{fig:stop_1}
    }
    \\
    \subfloat[receiving \textit{stop} instruction]{
        \includegraphics[width=1\textwidth]{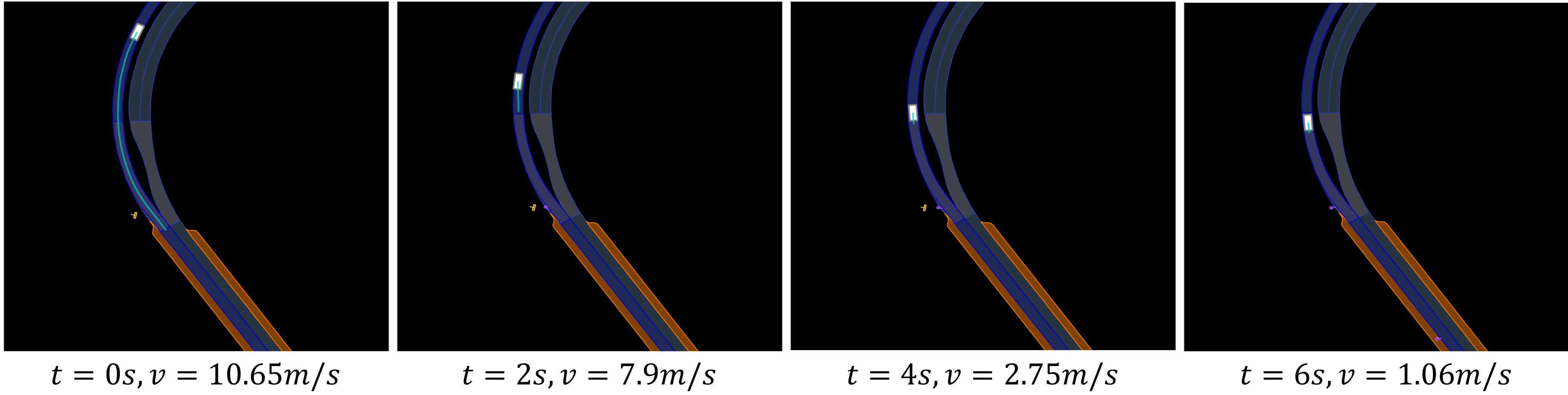}
        \label{fig:stop_2}
    }
    \caption{\textbf{Visualization of AsyncDriver following  human instruction.} The light blue line represents ego’s trajectory for the next $8$ seconds. It contrasts the planning trajectories of the ego receiving conventional routing instructions against a forced \textit{stop} instruction.}

    \label{fig:stop}
\end{figure}

\begin{table}[t]
  \caption{ \textbf{Replacing LLM with non-pretrained cross attention.} Utilize map information and learnable instruction embeddings to perform cross-attention, replacing the features generated by the LLM in the Scene-Associated Instruction Feature Extraction Module.
  }
  
  \centering
  \setlength{\tabcolsep}{3.5pt}
  \setlength{\arrayrulewidth}{0.2pt}
\begin{tabular}{ccc c}
    \toprule
    \makecell{Hidden Feature \\ Dimensions}      & \makecell{Instruction \\ Embedding} & LLM & Score \\
    \midrule
    256  & \checkmark &     & 63.27 \\
    5120   & \checkmark &    & 63.59  \\ 
    5120   &            & \checkmark    & \textbf{65.00}  \\
  \bottomrule
  \end{tabular}
  \label{tab:only_navi}
\end{table}

\subsection{Ablation Study}

In this section, we explore the necessity of LLM and investigate the effectiveness of individual components in AsyncDriver. 
\subsubsection{Necessity of LLM}
We conducted experiments by replacing the LLM with transformer blocks of different dimensions (256 and 5120), incorporating learnable routing instruction embeddings. Meanwhile, the Alignment Assistance Module and Adaptive Injection Block remained unchanged. The results, shown in 
Tab. \ref{tab:only_navi}, indicate that despite a 20-fold increase in the number of transformer parameters, the performance only marginally improved from 63.27 to 63.59. In contrast, our AsyncDriver achieved a performance of 65.00, highlighting the significant impact of pretrained knowledge from LLM.

\subsubsection{Ablation of Components}
Integrating a simple MLP as a prediction head after the LLM for planning significantly degraded the performance, indicating that simple trajectory regression cannot effectively align multi-modal information, thus failing to leverage the LLM's knowledge for scene reasoning. We replaced the MLP with a real-time planner and progressively added four structures: (i) Adaptive Injection Block, (ii) Alignment Assistance Module, (iii) LoRA, and (iv) pre-trained LoRA. As shown in Tab. \ref{tab:abl}, each module improved performance, with the Alignment Assistance Module and pre-trained LoRA weights contributing the most, yielding score increases of $0.94$ and $0.97$, respectively.

\begin{table}[tb]
  \caption{ \textbf{Component ablation study.}  MLP, RT-Planner, Ada, Align, LoRA, and LoRAPre represent i) direct regression of waypoints using MLP, ii) integration of a real-time planner, iii) Adaptive Injection Block, iv) Alignment Assistance Module, v) incorporation of LoRA, and vi) pre-trained LoRA with \textit{Reasoning1K}, respectively.  }
  
  \centering
  \setlength{\tabcolsep}{3.5pt}
  \setlength{\arrayrulewidth}{0.2pt}
\begin{tabular}{ccccccc}
\toprule
MLP & RT-Planner & Ada & Align & LoRA & LoRAPre & Score \\
\midrule
\checkmark   &           &     &           &      &         & 33.91 \\
    & \checkmark         &     &           &      &         & 62.01 \\
    & \checkmark         & \checkmark   &           &      &         & 62.84 \\
    & \checkmark         & \checkmark   & \checkmark         &      &         & 63.78 \\
    & \checkmark         & \checkmark   & \checkmark         & \checkmark    &         & 64.03 \\
    & \checkmark         & \checkmark   & \checkmark         & \checkmark    & \checkmark       & \textbf{65.00}  \\
  \bottomrule
\end{tabular}
  \label{tab:abl}
\end{table}

\section{Conclusions}
\label{sec:con}

In this paper, we introduce AsyncDriver, a new asynchronous LLM-enhanced, closed-loop framework for autonomous driving.  By aligning vectorized scene information with a series of routing instructions to form multi-modal features, we fully leverage LLM's capability for scene reasoning, extracting scene-associated instruction features as guidance. Through the proposed Adaptive Injection Block, we achieve the integration of series of routing information into any transformer-based real-time planner, enhancing its ability to understand and follow language instructions, and achieving outstanding closed-loop performance in nuPlan's challenging scenarios. Notably, owing to the structural design of our method, it supports asynchronous inference between the LLM and the real-time planner. Experiments show that our approach significantly increases inference speed with minimal loss in accuracy, reducing the computational costs introduced by LLM.

\subsubsection{Broader Impacts And Limitations.}

Should the method prove successful, the proposed asynchronous inference scheme could significantly enhance the prospects for integrating LLMs into the practical application within the autonomous driving sector. Nevertheless, while this research has employed LLMs, it falls short of substantiating their generalization properties for the planning task. Future endeavors aim to rigorously assess the generalization and transfer potential of LLMs in in vectorized scenarios.

\newpage

\section*{Acknowledgements}
This research is supported in part by the National Science and Technology Major Project (No. 2022ZD0115502), the National Natural Science Foundation of China (No. 62122010, U23B2010), Zhejiang Provincial Natural Science Foundation of China (No. LDT23F02022F02), Beijing Natural Science Foundation (No. L231011), Beihang World TOP University Cooperation Program, and Lenovo Research.
\newpage
\appendix

\renewcommand{\thefigure}{S\arabic{figure}}

\section*{Appendix}
In this supplementary material, we first present the templates for QAs and prompts of pre-training and fine-tuning data (Appendix \ref{sec:sup_prompt}). Then, we provide detailed metric information for our closed-loop evaluation (Appendix \ref{sec:sup_metric}), as well as the specific 14 scenario types defined in nuPlan challenges \cite{nuplanchallange} (Appendix \ref{sec:sup_types}). In the end, we report additional quantitative results conducted to further substantiate the efficacy of our approach (Appendix \ref{sec:sup_vis}).

\section{Data Templates}
\label{sec:sup_prompt}
\subsection{Pre-training Data}
\subsubsection{Planning-QA}
We assessed the capabilities of Llama2-13B \cite{touvron2023llama} via a zero-shot methodology and discovered that its extensive pre-training data provides a solid foundation in traffic rule comprehension. However, its limited mathematical prowess poses challenges in grasping and deducing the connections between instructions and numerical expressions. To address this, we introduced a language-based \textit{Planning-QA} dataset aimed at transitioning LLM from a general model to a specialized model adept in autonomous driving planning. This enhancement focuses on refining its capability in instruction interpretation and reasoning. 

Concretely, we delineated the level of autonomous driving planning into three granularities: 1) high-level instructions: formulated through velocity commands including \textit{stop}, \textit{accelerate}, \textit{decelerate}, \textit{maintain speed}, and routing commands including \textit{turn left}, \textit{turn right}, \textit{go straight}, 2) control: assessing the values of velocity and acceleration, 3) and waypoint: encompassing a series of points. Six question types were devised to articulate the transitional relationships across the \textit{high-level instructions - control - waypoint} spectrum, and each QA-pair is adapted based on log data from nuPlan \cite{caesar2021nuplan}. Fig. \ref{fig:templet_1_short} illustrates the universal system prompt template applicable to all questions, whereas Fig. \ref{fig:Q1}-\ref{fig:Q6} display specific examples for each question type alongside their respective answer, substituting \textbf{<Question>} and \textbf{<Answer>} within the system prompt.

\begin{figure}[htbp]
    \setcounter{figure}{0}
    \centering
    \captionsetup[subfloat]{font=scriptsize, labelfont=normalfont, textfont=normalfont}
    \subfloat[system prompt template for Planning-QA dataset]{
        \includegraphics[width=1\textwidth]{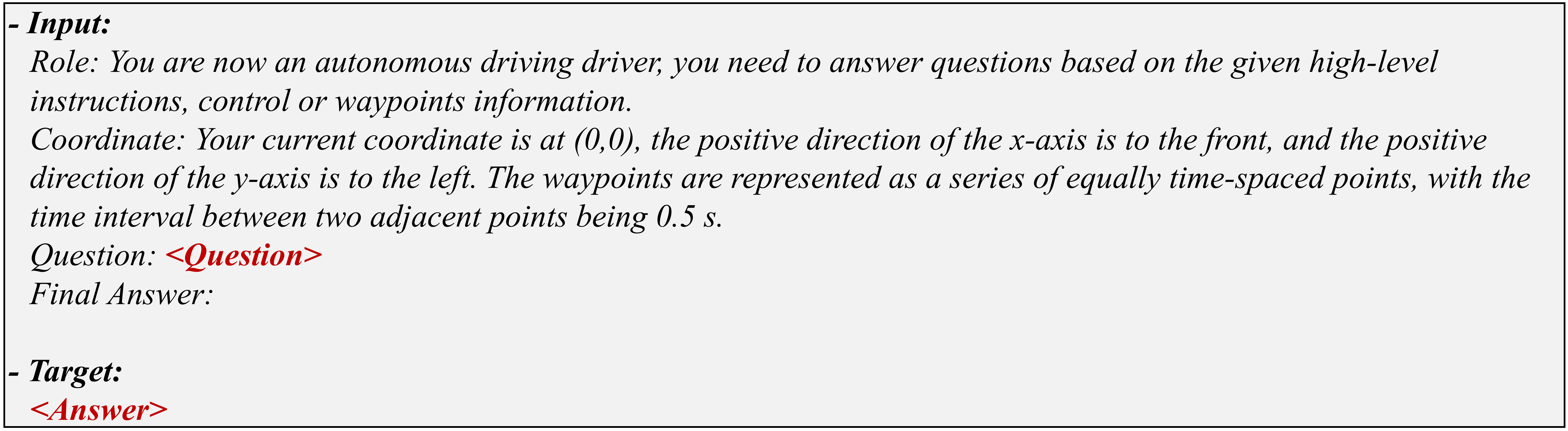}
        \label{fig:templet_1_short}
    }
    \vspace{-5mm}
\end{figure}

\begin{figure}[tbp]
    \ContinuedFloat
    \centering
    \captionsetup[subfloat]{font=scriptsize, labelfont=normalfont, textfont=normalfont}
    \subfloat[\textit{waypoints - high-level instructions}]{
        \includegraphics[width=1\textwidth]{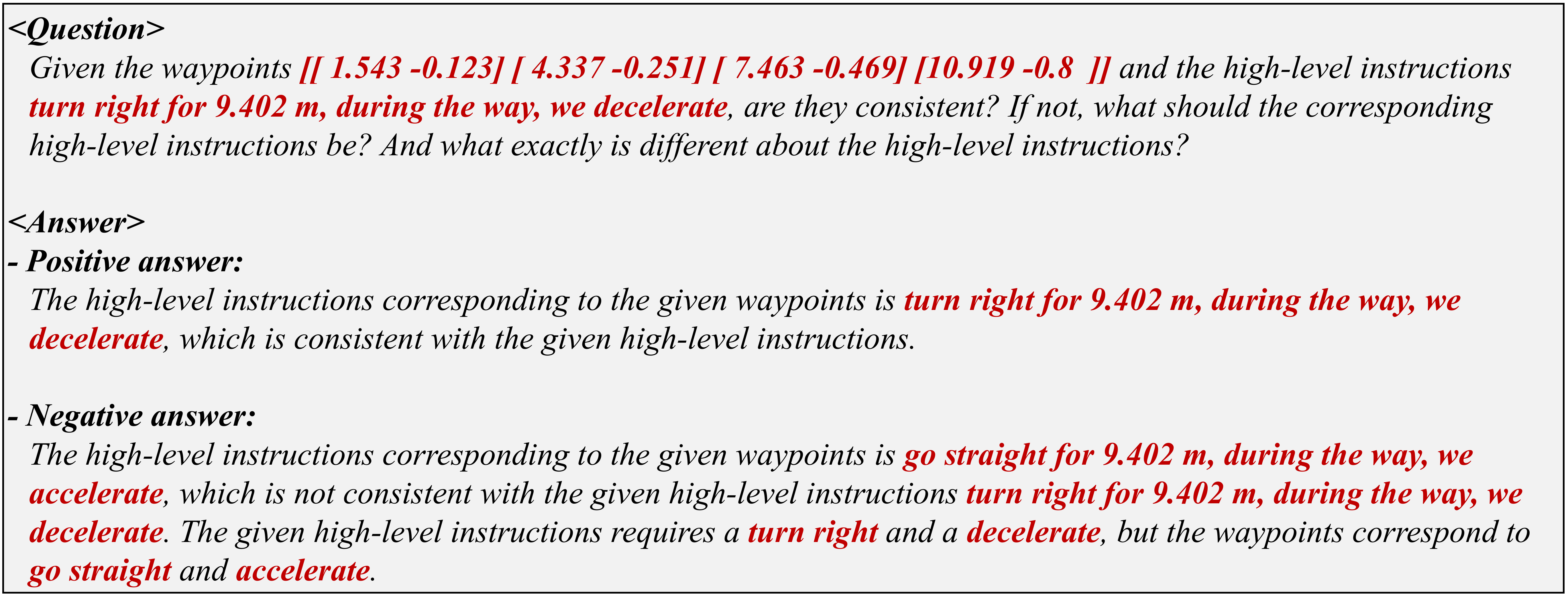}
        \label{fig:Q1}
    }
    \\
    \subfloat[\textit{waypoints - control}]{
        \includegraphics[width=1\textwidth]{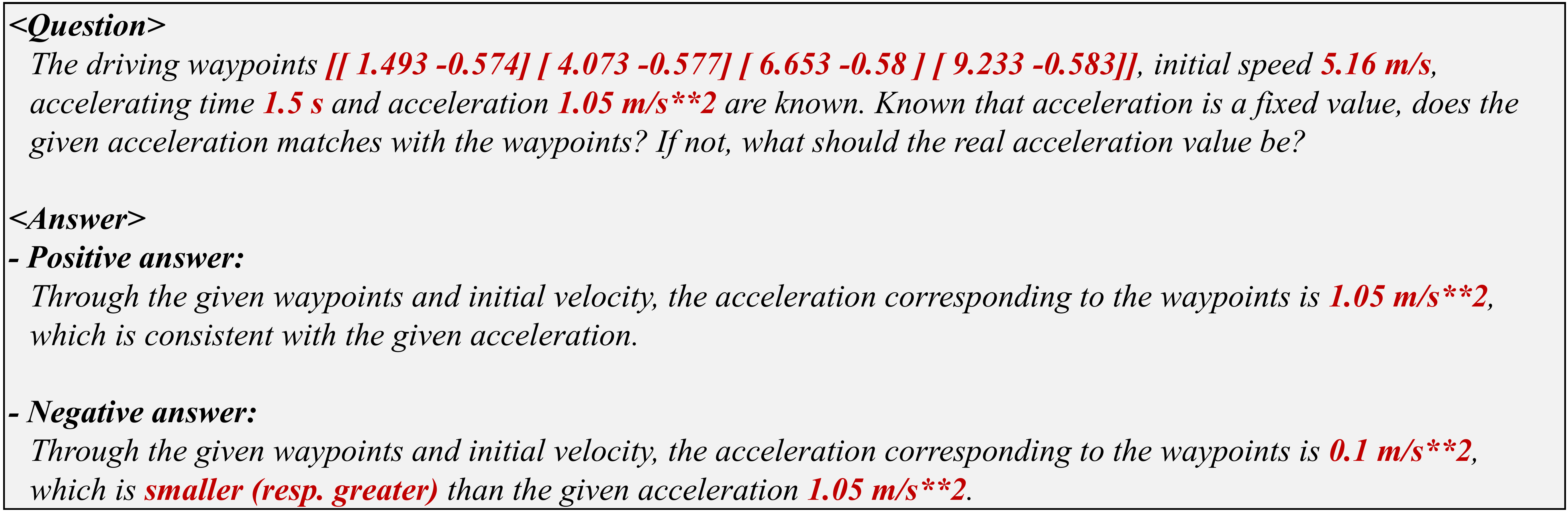}
        \label{fig:Q2}
    }
    \\
    \subfloat[\textit{high-level instructions - waypoint}]{
        \includegraphics[width=1\textwidth]{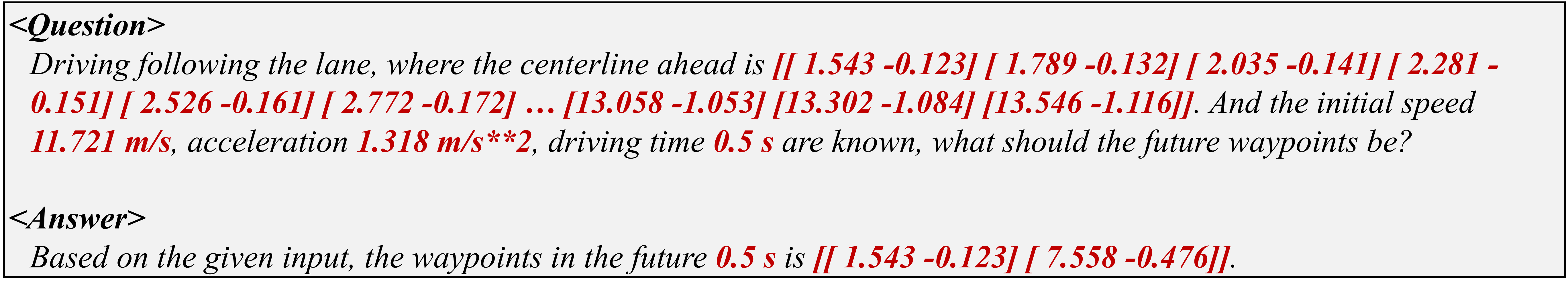}
        \label{fig:Q3}
    }
    \\
    \subfloat[\textit{high-level instructions - control}]{
        \includegraphics[width=1\textwidth]{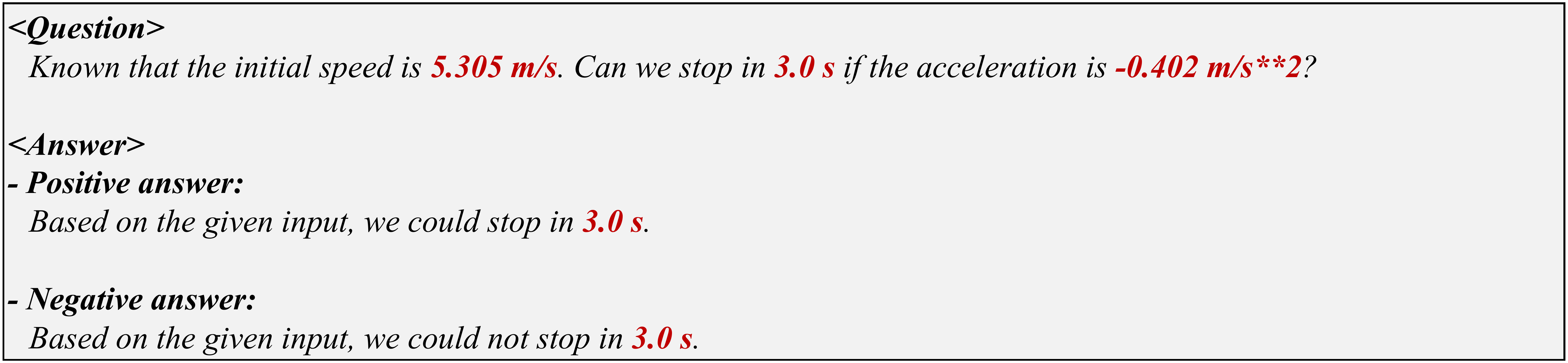}
        \label{fig:Q4}
    }
    \\
    \subfloat[\textit{control - waypoint}]{
        \includegraphics[width=1\textwidth]{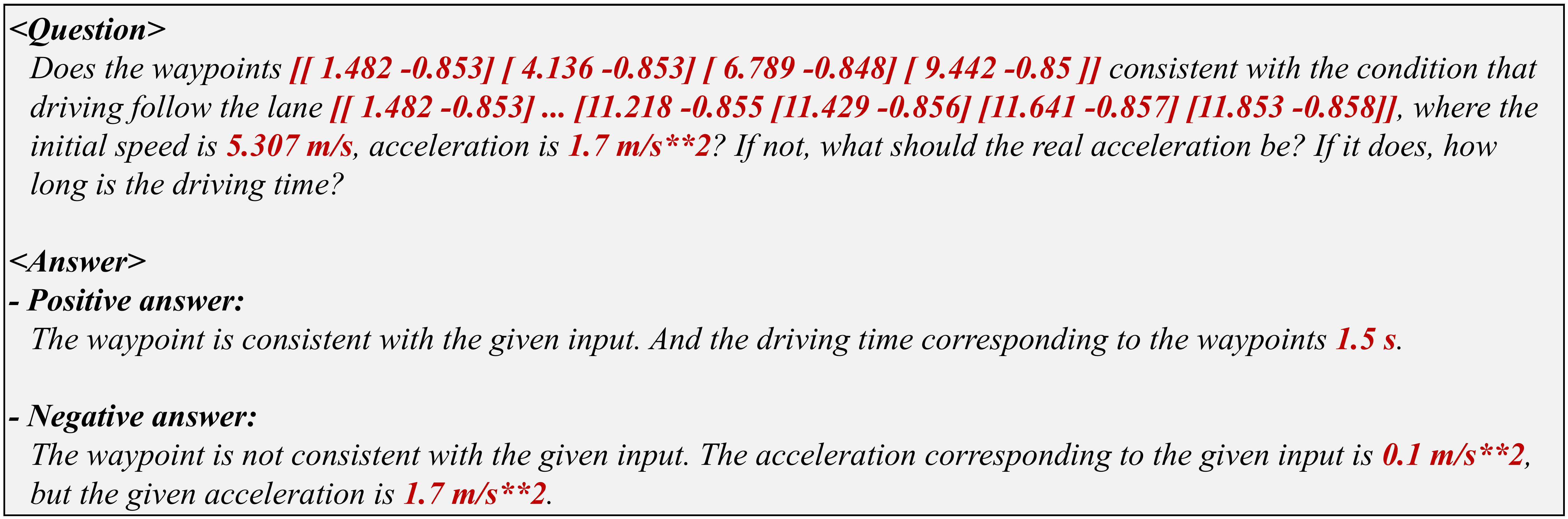}
        \label{fig:Q5}
    }
    \vspace{-5mm}
    \label{fig:s1bf}
\end{figure}

\begin{figure}[tbp]
    \ContinuedFloat
    \centering
    \captionsetup[subfloat]{font=scriptsize, labelfont=normalfont, textfont=normalfont}
    \subfloat[\textit{control - high-level instructions}]{
        \includegraphics[width=1\textwidth]{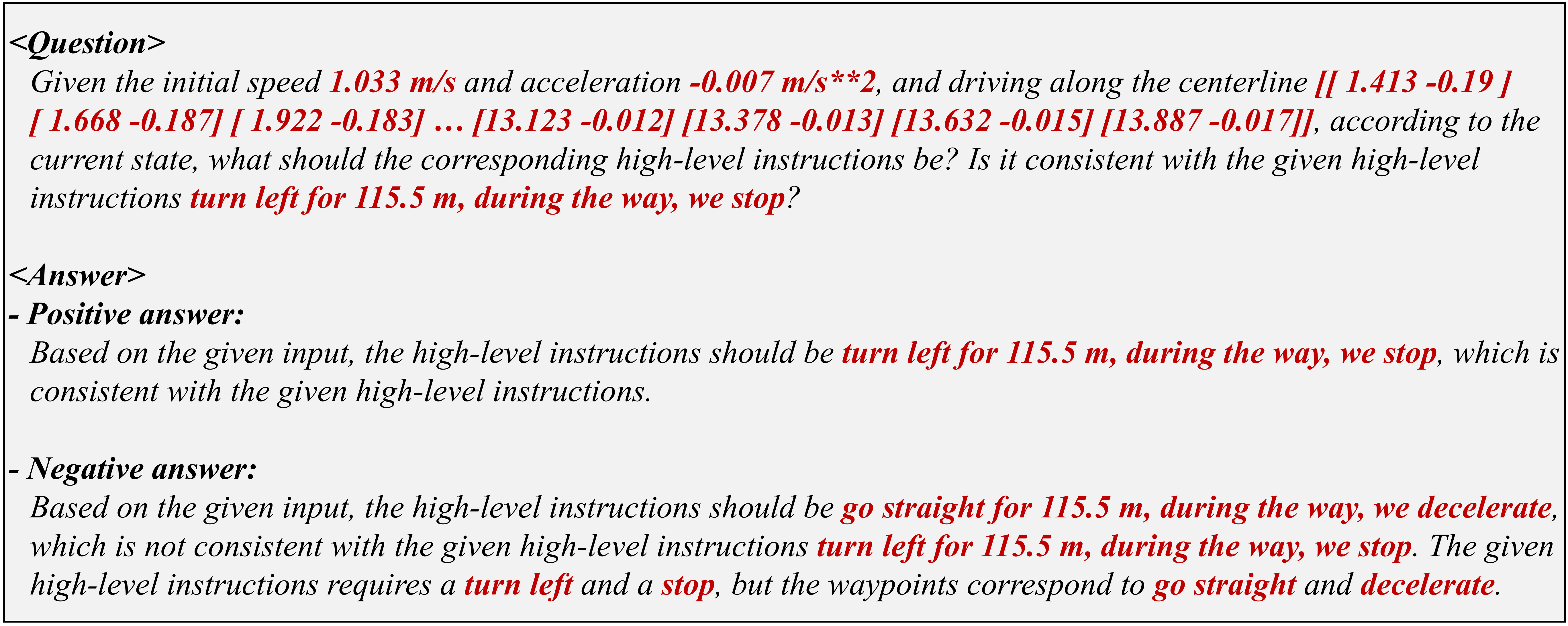}
        \label{fig:Q6}
    }
    \caption{Six question types were devised to articulate the transitional relationships across the \textit{high-level instructions - control - waypoints} spectrum. All data are generated based on logs from nuPlan, with the red parts being subject to change based on actual data. Some data have both positive and negative answers, which are both displayed in the corresponding templates. The waypoints are abbreviated due to space limitations.}
    \label{fig:S1}
    \vspace{-5mm}
\end{figure}

\subsubsection{Reasoning1K}
\begin{figure}[tbp]
    \centering
\captionsetup[subfloat]{font=scriptsize, labelfont=normalfont, textfont=normalfont}
    \subfloat[example of input for Reasoning1K]{
        \includegraphics[width=1\textwidth]{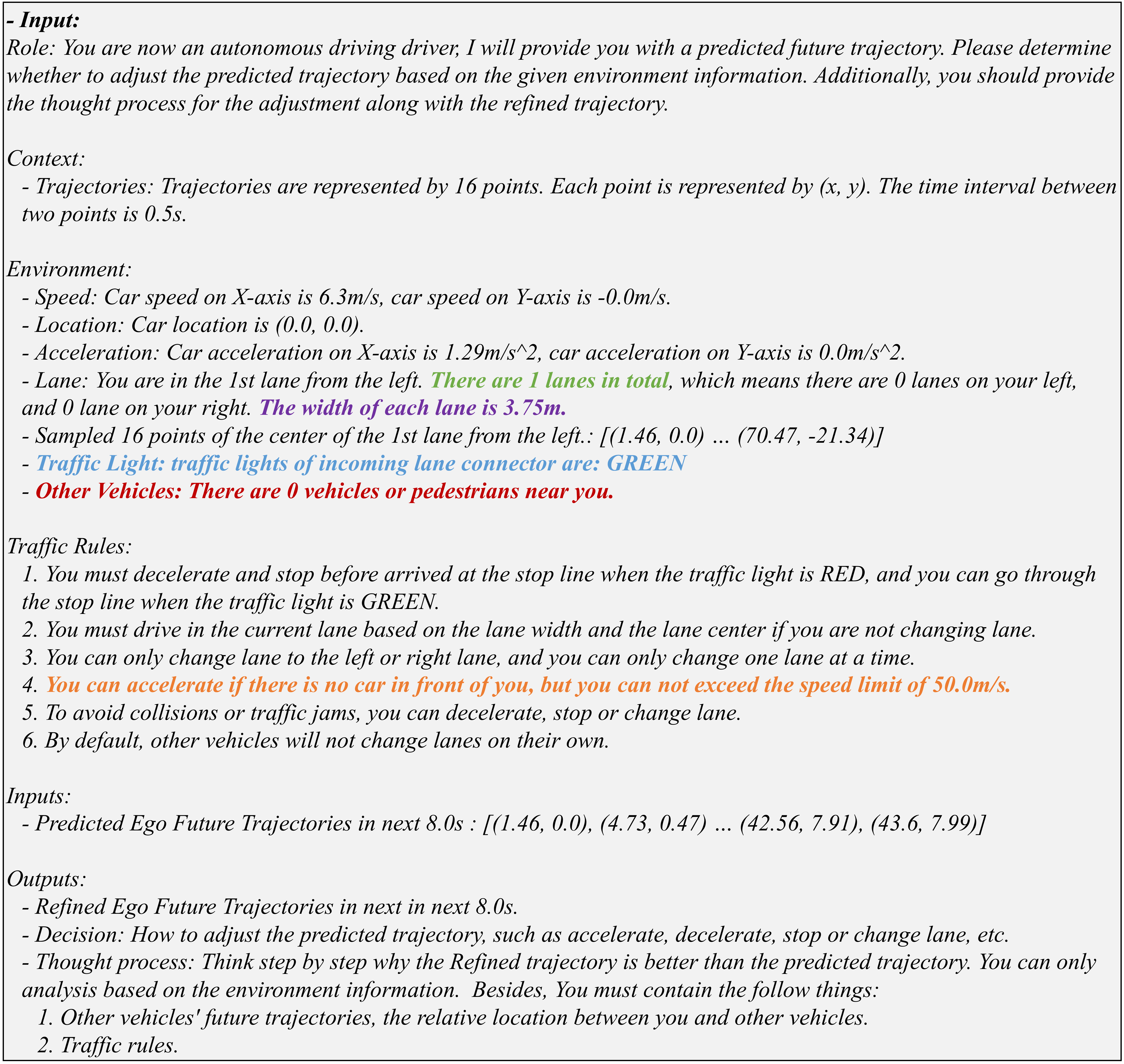}
        \label{fig:reasoning_input}
    }
    \\
    \subfloat[example of corresponding target for Reasoning1K]{
        \includegraphics[width=1\textwidth]{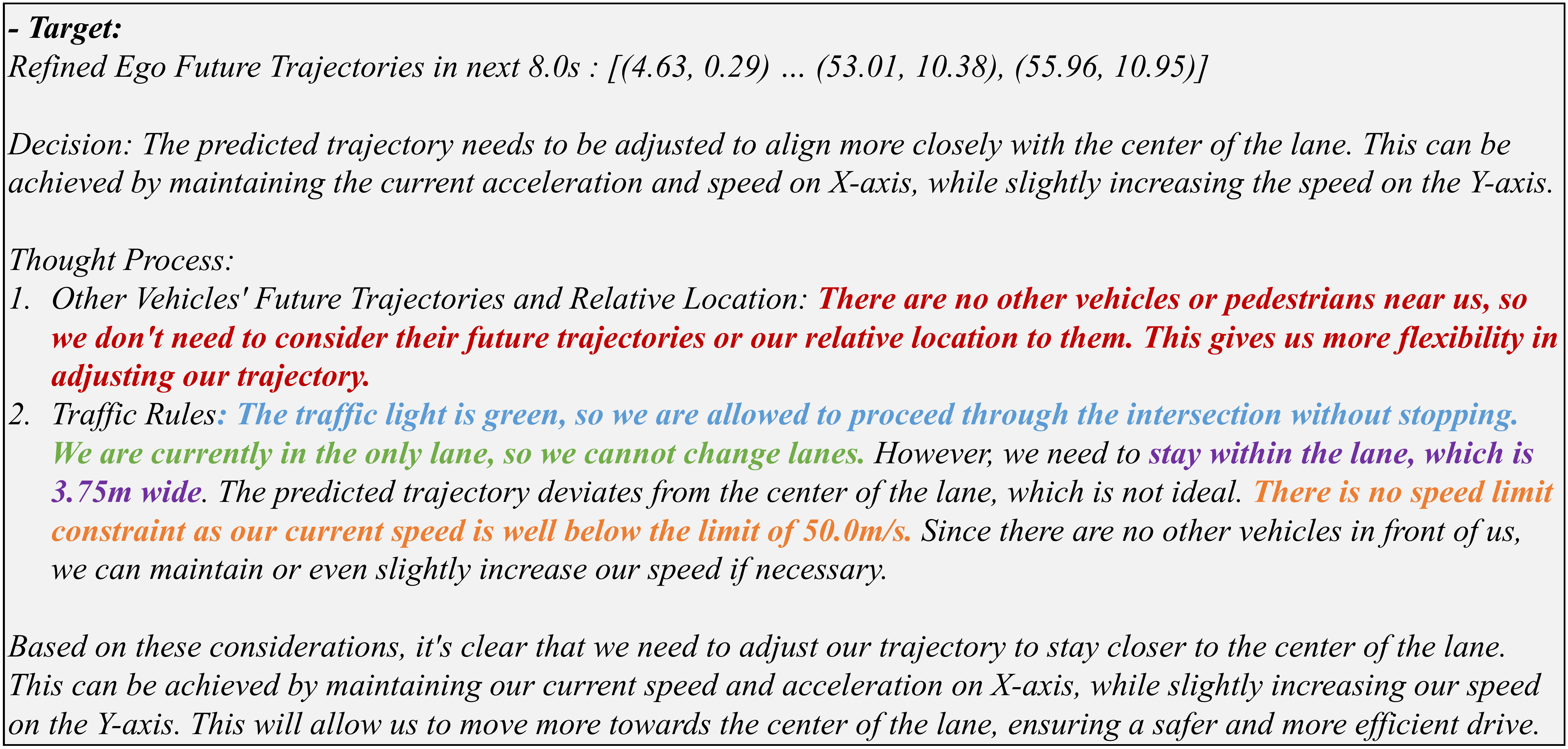}
        \label{fig:reasoning_target}
    }
    \caption{An example from \textit{Reasoning1K}, where \ref{fig:reasoning_input} displays the input, and \ref{fig:reasoning_target} shows the corresponding target. The reasoning-related parts are highlighted in color, with the same color in both input and target representing the cause and effect in the reasoning process respectively.}
    \vspace{-5mm}
\end{figure}

Additionally, we have crafted 1,000 QA-pairs using GPT-4, tailored to tasks that require not just decision-making but also in-depth reasoning. Each sample includes a question that sets the context, provides scene descriptions, outlines considerations based on traffic rules, details the anchor trajectory from real-time planners, and specifies requirements for the answers, while the answer delivers trajectory predictions along with the decisions and the underlying thought process. By training with mixed of \textit{Reasoning1K} and \textit{Planning-QA}, we have substantially enriched the dataset's diversity and complexity. This approach aims to bolster LLM's ability to deduce answers through logical reasoning, seamlessly integrating traffic rules with detailed, scenario-based analysis. Fig. \ref{fig:reasoning_input} and Fig. \ref{fig:reasoning_target} showcase an example of the input and its corresponding target from the dataset, respectively.

\subsection{Fine-tuning Data}
The pre-training phase aims to utilize language-based data to bolster the LLM's understanding of instructions, whereas the fine-tuning phase seeks to further refine this comprehension by aligning it with vectorized scenes. This entails a process characterized by multi-modal inputs and vectorized supervision. Inputs encompass two parts: linguistic prompts and vectorized scene data. The linguistic portion includes fixed system prompts and log-based routing instructions. Within this template, the \textbf{<map>} token is substituted with tokens of vectorized scene at the point of input. Instructions are generated from the forthcoming 8 seconds' logs through a hand-crafted rules, encompassing four instruction categories—\textit{turn left}, \textit{turn right}, \textit{go straight}, and \textit{stop} — paired with pertinent distance details. Notably, the \textit{stop} is only employed when the actual path length for the next 8 seconds falls below 0.5 meters. An illustration of the input template is provided in Fig. \ref{fig:finetuning_prompt}.

\begin{figure}[tbp]
    \centering
    \includegraphics[width=1\textwidth]{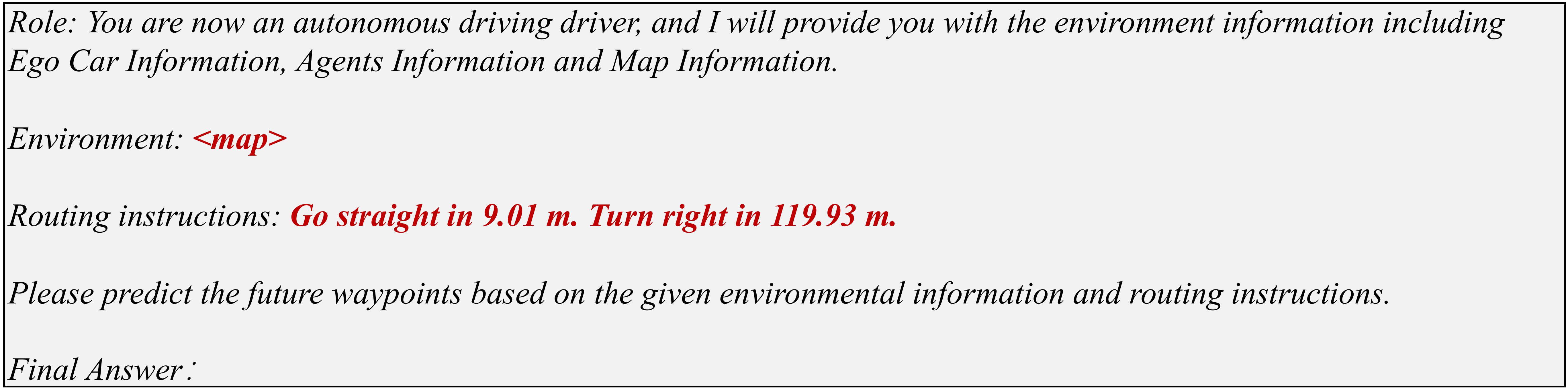}
    \caption{Prompt template for fine-tuning stage.}
    \label{fig:finetuning_prompt}
    \vspace{-0.5cm}
\end{figure}

\section{Closed-loop Metrics}
\label{sec:sup_metric}
In this section, we elucidate the metrics of closed-loop evaluation deployed within the nuPlan \cite{caesar2021nuplan} framework. The calculation for the composite metric \textit{Score} is detailed in Formula \ref{for:score}.

\noindent \textbf{Drivable Area Compliance (Drivable)} Ego should remain within the designated drivable area. If a frame is identified where the maximum distance from any corner of the ego's bounding box to the nearest drivable area exceeds a predetermined threshold, the drivable area compliance score is assigned a value of $0$; otherwise, it is assigned a value of $1$.

\noindent \textbf{Driving Direction Compliance (Direct.)} The evaluation of the ego's compliance with the correct driving direction is encapsulated within this metric. It stipulates that a distance traversed in the opposite direction of less than $2$ meters within a 1-second interval is deemed compliant, warranting a score of $1$. A traversal exceeding $6$ meters within the same timeframe indicates a significant deviation from compliance, thus yielding a score of $0$. Intermediate distances are assigned a proportional score of $0.5$.

\noindent \textbf{Ego is Comfortable (Comf.)} This metric quantifies the comfort of the trajectory based on a series of kinematic and dynamic thresholds including longitudinal acceleration, lateral acceleration, yaw acceleration, yaw velocity, longitudinal jerk, and overall jerk magnitude. Compliance across all specified parameters results in a trajectory being classified as "comfortable", meriting a score of $1$; otherwise, the score is $0$.

\noindent \textbf{Ego Progress Along Expert Route (Prog.)} This metric represents the ratio of the ego‘s progress per frame to that of an expertly defined route. The closer the ego's progress is to the expert route's progress, the higher the score, with the maximum score being $1$.

\noindent \textbf{No Ego At-fault Collisions (Coll.)} This metric is integral to the safety evaluation, quantifying instances where the ego is directly responsible for collisions. An absence of at-fault collisions throughout the evaluation period scores a $1$. A singular collision with a static object, such as a traffic cone, is deemed a minor infraction, scoring a $0.5$, whereas all other scenarios result in a score of $0$, accentuating the paramount importance of collision avoidance.

\noindent \textbf{Speed Limit Compliance (Lim.)} The ego's adherence to speed regulations is scrutinized by calculating the average instance of speed limit violations per frame. This metric deducts points based on the proximity of the violation to the maximum speed threshold, promoting strict compliance with speed limits.

\noindent \textbf{Time to Collision within Bound (TTC)} This metric evaluates the hypothetical time to collision (TTC) with any external agent if the ego were to continue on its trajectory with unaltered speed and heading. A TTC exceeding $0.95$ seconds is considered safe, thus earning a score of $1$. Lower TTC values signify elevated risk, consequently scoring a $0$, highlighting the significance of maintaining a safe buffer from other road users.

\noindent \textbf{Making Progress (MP)} An ancillary metric to the Ego Progress Along Expert Route, it assigns a score of $1$ if the progress exceeds a threshold of $0.2$, otherwise scoring $0$. Though not directly incorporated into the main text due to its derivative nature, it contributes to the final score.

\noindent \textbf{Score} The final scenario score is calculated by combining the above metrics in a predetermined manner, providing a comprehensive assessment of the autonomous vehicle's performance within the evaluated scenarios:

\begin{align}
Score= (Prog*5+TTC*5+Lim*4+Comf*2) \div 16 \notag \\ *Coll*Drivable*MP*Direct
\label{for:score}
\end{align}
\section{Scenario Types}
\label{sec:sup_types}

In this section, we delineate $14$ scenario types employed in both the training and evaluation which is consistent with the nuPlan Challenge 2023 \cite{nuplanchallange}. These scenarios encompass a broad spectrum of driving conditions to evaluate the adaptability and planning capabilities of autonomous vehicles. The scenarios are categorized as follows:

\noindent \textbf{Behind Long Vehicle (type 0)}: The ego vehicle is positioned behind a long vehicle, maintaining a longitudinal distance of $3$ to $10$ meters within the same lane, where the lateral distance is less than $0.5$ meters.

\noindent \textbf{Changing Lane (type 1)}: Initiates a maneuver to transition towards an adjacent lane at the beginning of the scenario.

\noindent \textbf{Following Lane with Lead (type 2)}: Involves the ego vehicle following a leading vehicle that is moving in the same lane with a velocity exceeding $3.5\, \text{m/s}$ and a longitudinal distance of less than $7.5$ meters.

\noindent \textbf{High Lateral Acceleration (type 3)}: Characterized by the ego vehicle experiencing high acceleration ($1.5 < \text{acceleration} < 3\, \text{m/s}^2$) across the lateral axis, accompanied by a high yaw rate, without executing a turn.

\noindent \textbf{High Magnitude Speed (type 4)}: The ego vehicle achieves a high velocity magnitude exceeding $9\, \text{m/s}$ with low acceleration.

\noindent \textbf{Low Magnitude Speed (type 5)}: Ego vehicle moves at low velocity magnitude ($0.3 < \text{velocity} < 1.2\, \text{m/s}$) with low acceleration, without coming to a stop.

\noindent \textbf{Near Multiple Vehicles (type 6)}: The ego vehicle is proximate to multiple (more than six) moving vehicles within a distance of less than $8$ meters while maintaining a velocity of more than $6\, \text{m/s}$.

\noindent \textbf{Starting Left Turn (type 7)}: Marks the commencement of a left turn by the ego vehicle across an intersection area, without being halted.

\noindent \textbf{Starting Right Turn (type 8)}: Marks the commencement of a right turn by the ego vehicle across an intersection area, without being halted.

\noindent \textbf{Starting Straight Traffic Light Intersection Traversal (type 9)}: The ego vehicle begins to traverse straight through an intersection controlled by traffic lights, without being stopped.

\noindent \textbf{Stationary in Traffic (type 10)}: The ego vehicle remains stationary amidst traffic, surrounded by multiple (more than six) vehicles within an $8$-meter radius.

\noindent \textbf{Stopping with Lead (type 11)}: The scenario begins with the ego vehicle initiating deceleration (acceleration magnitude less than $-0.6\, \text{m/s}^2$, velocity magnitude less than $0.3\, \text{m/s}$) due to the presence of a leading vehicle ahead within a distance of less than $6$ meters.

\noindent \textbf{Traversing Pickup/Dropoff (type 12)}: The ego vehicle navigates through a pickup or drop-off area without stopping.

\noindent \textbf{Waiting for Pedestrian to Cross (type 13)}: The ego vehicle waits for a pedestrian, who is within an $8$-meter distance and less than $1.5$ seconds away from the intersection, to cross a crosswalk. This occurs while the ego vehicle is not stopped and the pedestrian is not located in a pickup or drop-off area.

\section{Visualization}
\label{sec:sup_vis}
Fig. \ref{fig:s4} additionally showcases a series of our visualization results and demonstrates its superior performance to GameFormer \cite{huang2023gameformer} across multiple metrics.

\begin{figure}[htbp]
    \ContinuedFloat
    \centering
\captionsetup[subfloat]{font=scriptsize, labelfont=normalfont, textfont=normalfont}
    \subfloat[GameFormer collides with other vehicles at 6.5s; AsyncDriver first yield to the vehicle going straight, then passes smoothly.]{
        \includegraphics[width=1\textwidth]{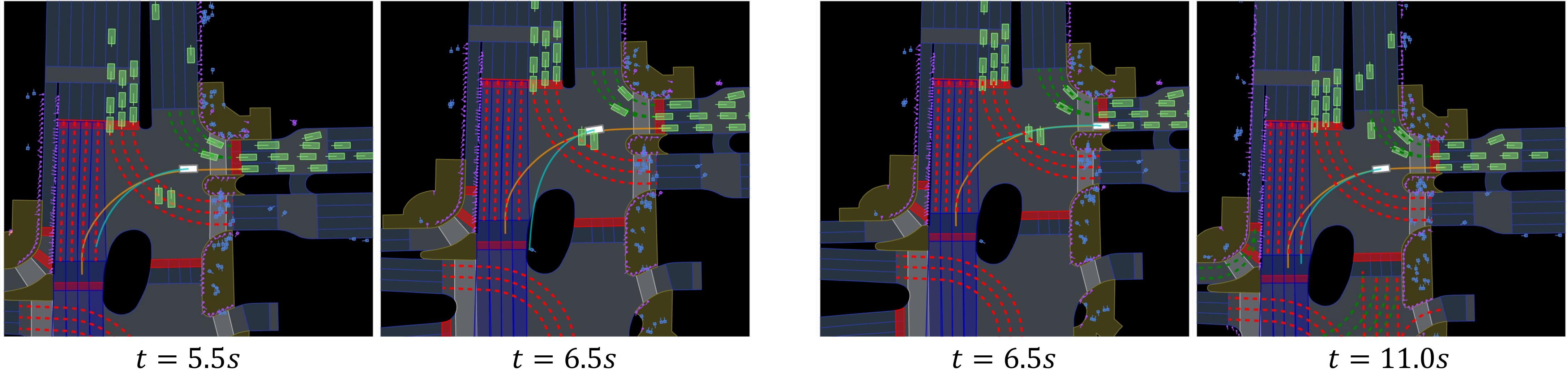}
        \label{fig:col_1}
    }
    \\
    \subfloat[GameFormer, upon reaching 11.5s, finds it difficult to evade pedestrians ahead, resulting in a collision at 15.0s; AsyncDriver, anticipating the situation in advance, passes smoothly before the pedestrians approach.]{
        \includegraphics[width=1\textwidth]{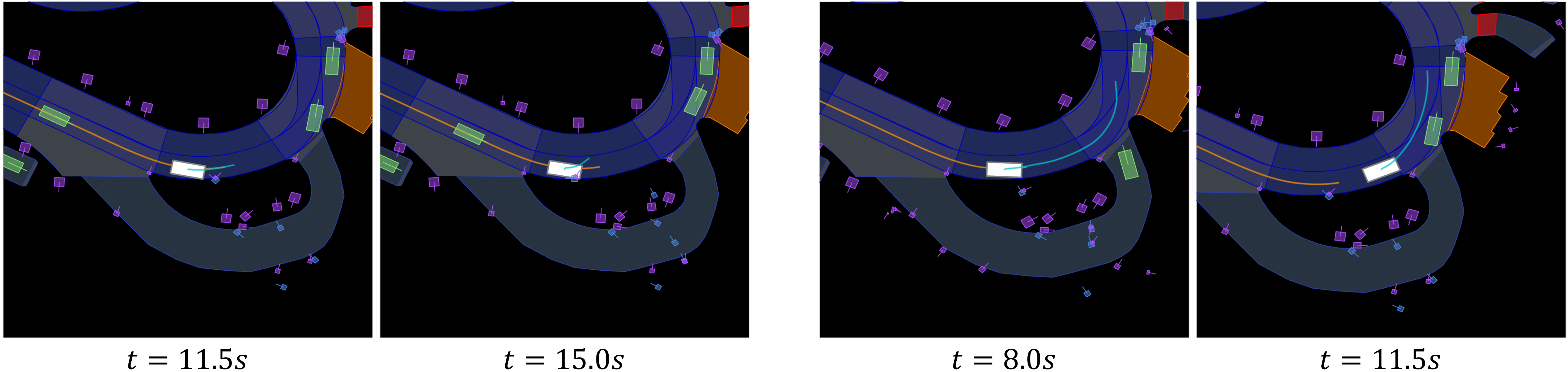}
        \label{fig:col_2}
    }
    \vspace{-5mm}
    \label{fig:planningqa_1}
\end{figure}

\begin{figure}[htbp]
    \ContinuedFloat
    \centering
\captionsetup[subfloat]{font=scriptsize, labelfont=normalfont, textfont=normalfont}
    \subfloat[GameFormer makes a misjudgment about the traffic light status at 5.0s, noticeably decelerating, leading to its failure to enter the intersection at 12.5s due to slow progress; AsyncDriver maintains a steady speed and enters the intersection smoothly.]{
        \includegraphics[width=1\textwidth]{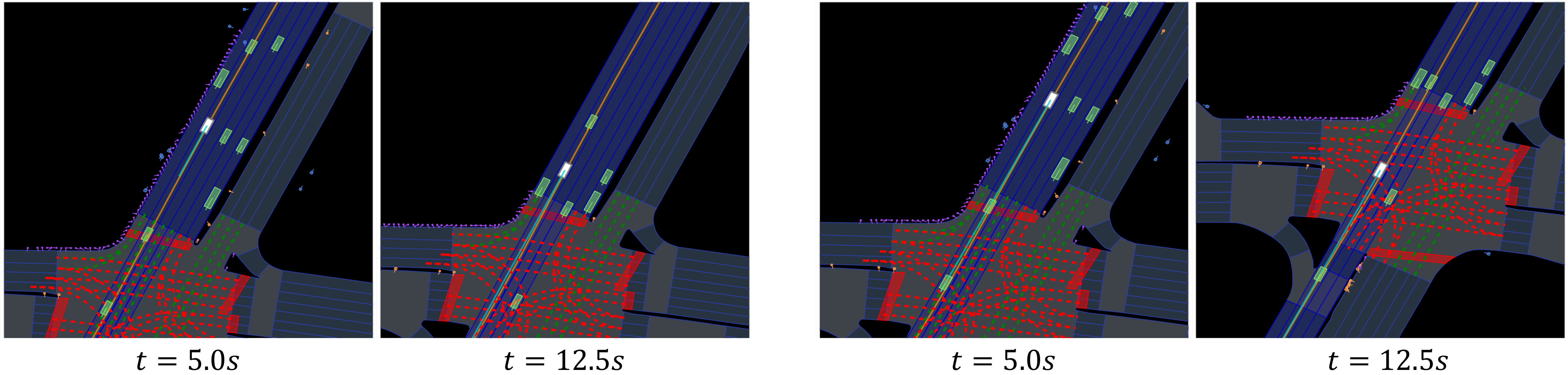}
        \label{fig:prog_1}
    }
    \\
    \subfloat[GameFormer, after the traffic light turns from red to green at 11.0s, fails to start promptly and remains stationary at 15.0s; AsyncDriver starts on time and enters the intersection.]{
        \includegraphics[width=1\textwidth]{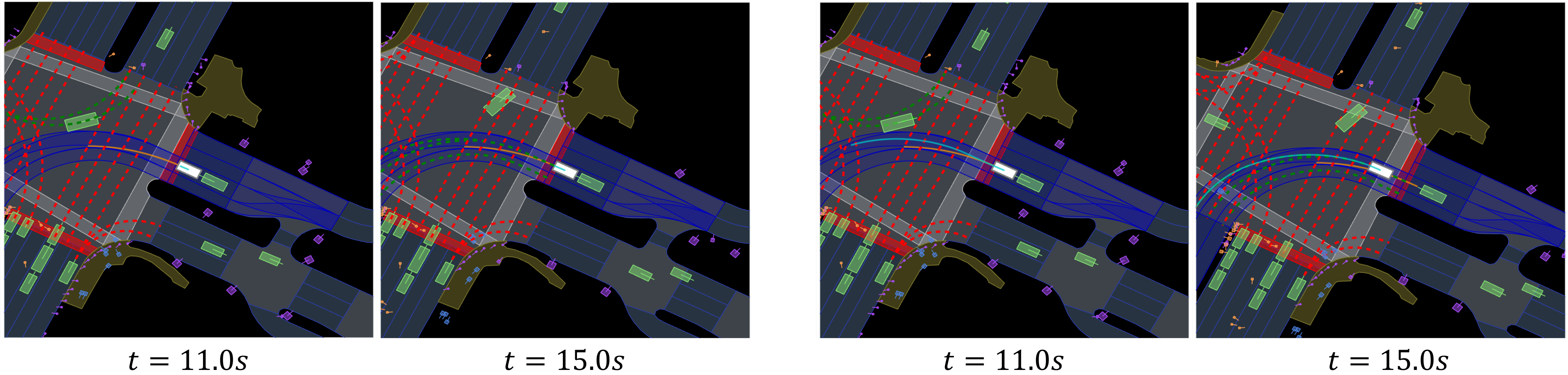}
        \label{fig:prog_2}
    }
    \\
    \subfloat[At 12.5 seconds, when a pedestrian appears ahead, GameFormer does not evade or decelerate; AsyncDriver clearly brakes.]{
        \includegraphics[width=1\textwidth]{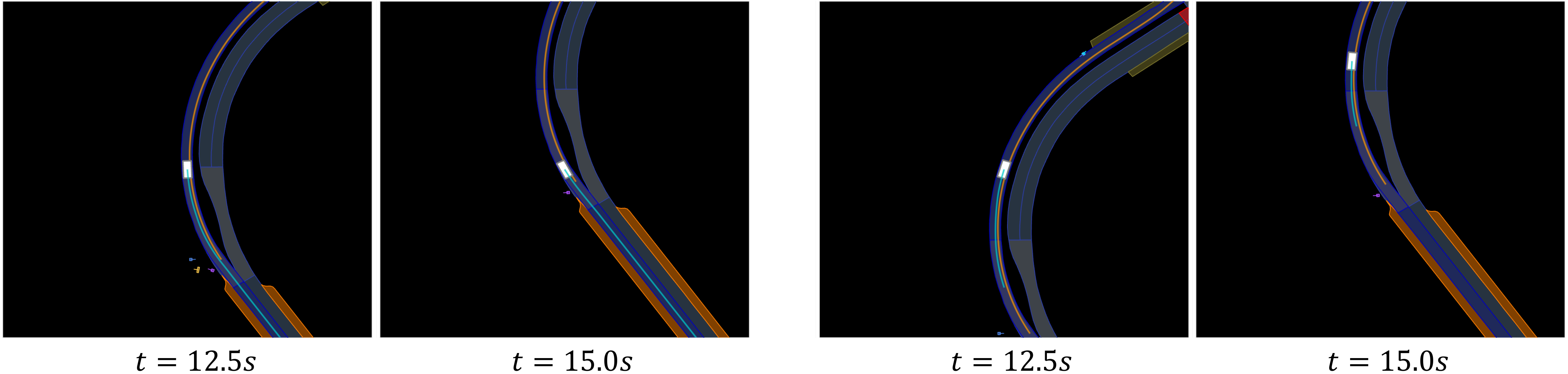}
        \label{fig:ttc_1}
    }
    \\
    \subfloat[GameFormer gets too close to the adjacent car while turning, increasing the risk of collision; AsyncDriver travels almost parallel to the neighboring car, greatly enhancing safety.]{
        \includegraphics[width=1\textwidth]{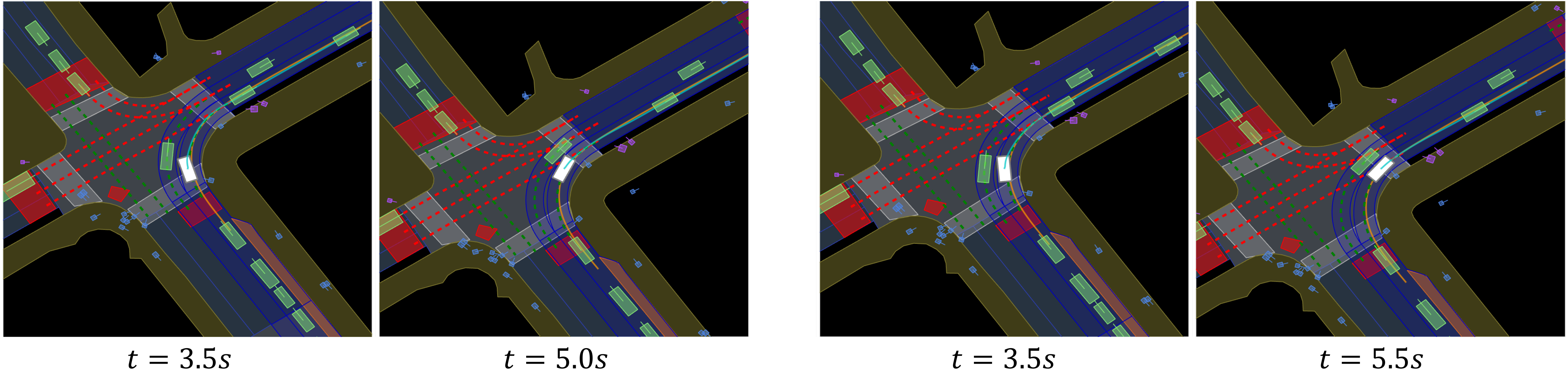}
        \label{fig:ttc_2}
    }
    \\
    \subfloat[GameFormer turns out of the drivable area; AsyncDriver does not.]{
        \includegraphics[width=1\textwidth]{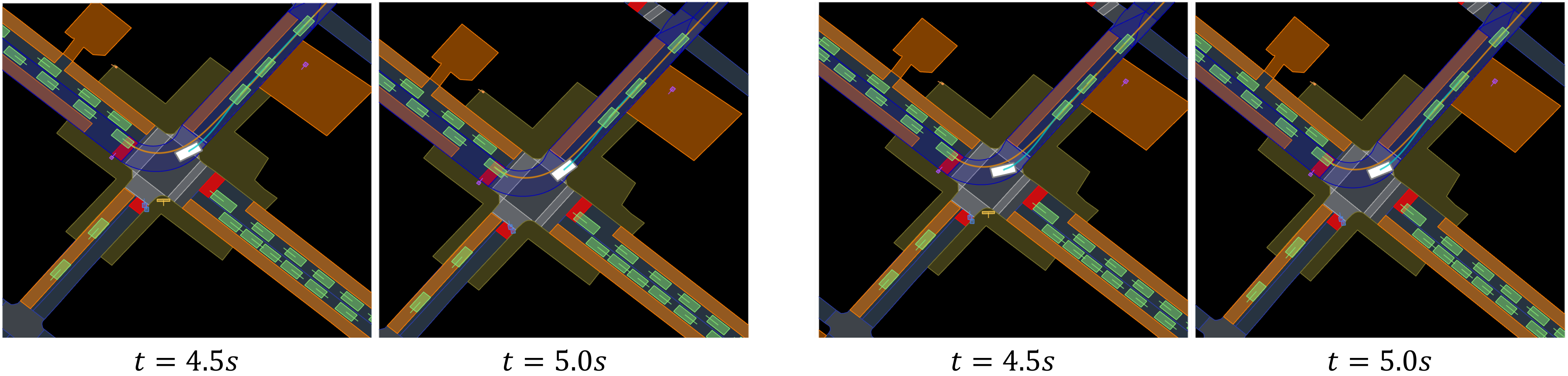}
        \label{fig:drivable_1}
    }
    \vspace{-5mm}
    \label{fig:planningqa_1}
\end{figure}

\begin{figure}[htbp]
    \ContinuedFloat
    \centering
    \setcounter{figure}{4}
\captionsetup[subfloat]{font=scriptsize, labelfont=normalfont, textfont=normalfont}
    \subfloat[GameFormer exceeds the drivable area; AsyncDriver does not.]{
        \includegraphics[width=1\textwidth]{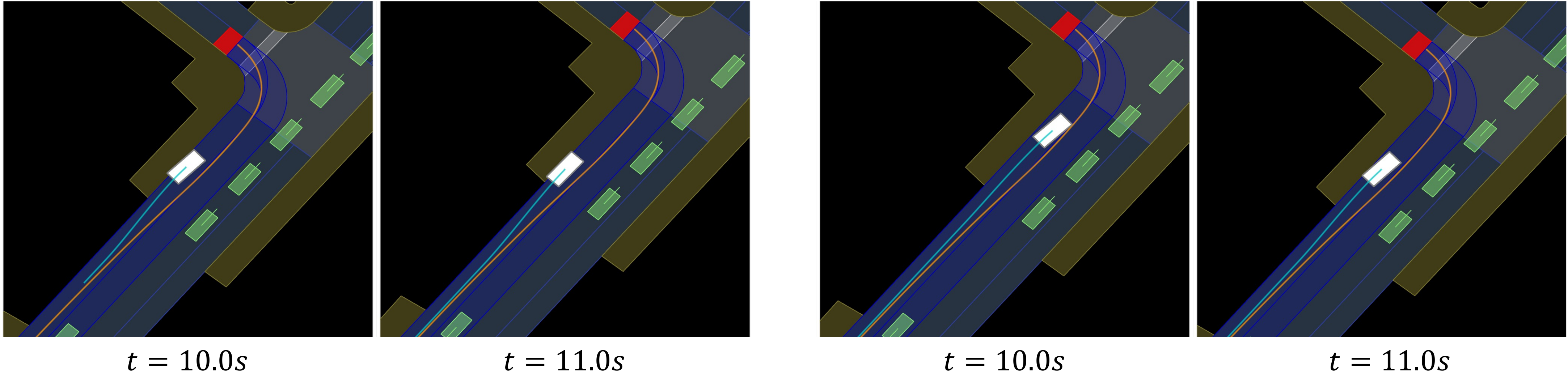}
        \label{fig:drivable_2}
    }
    \caption{The visualization results of GameFormer and AsyncDriver in various scenarios, with corresponding analytical explanations provided in subcaptions. The left two columns display the results for GameFormer, while the right two columns show the results for AsyncDriver.}
    \vspace{-5mm}
    \label{fig:s4}
\end{figure}

%
%
\newpage
\bibliographystyle{splncs04}
\bibliography{main}

\end{document}